\documentclass[runningheads]{llncs}

\usepackage{eccv}

\usepackage{eccvabbrv}

\usepackage{graphicx}
\usepackage{booktabs}
\usepackage{multirow}
\usepackage{bbding}
\usepackage{xcolor}
\usepackage{xspace}
\usepackage{colortbl}

\usepackage{amsmath}
\usepackage{amssymb}

\newcommand{\mName}{PGE-SAM\xspace}
\newcommand{\dataName}{DM-Seg}

\usepackage[pagebackref]{hyperref}

\begin{document}

\title{PGE-SAM: Prompt-Guided Feature Enhancement for Interactive Segmentation \\ under Degradation}

\author{Tuan-Duc Nguyen\inst{1} \and
Anh-Tuan Mai\inst{1} \and
Duc-Trong Le\inst{2}}

\authorrunning{T.~Nguyen et al.}

\institute{FPT Software AI Center \and
VNU University of Engineering and Technology \\
\email{trongld@vnu.edu.vn}}

\maketitle
\begin{abstract}
The Segment Anything Model (SAM) has revolutionized promptable image segmentation with strong zero-shot generalization. However, its performance degrades substantially under real-world imaging artifacts such as noise, blur, and compression. Existing methods restore features globally without focusing on segmentation-relevant regions and neglect SAM's iterative refinement mechanism, leading to suboptimal performance in interactive settings. We propose Prompt-Guided Feature Enhancement SAM (PGE-SAM), a framework that explicitly leverages user prompts and prior mask predictions to spatially guide the feature restoration process toward regions of interest through a Prompt Guidance Generator. To recover fine-grained details lost under degradation, we introduce Multi-Scale Features Interaction to incorporate low-level encoder features, along with a Foreground Reconstruction Loss that restricts feature-level supervision to the segmentation target. Furthermore, we present DM-Seg, a benchmark for interactive segmentation on degraded medical images, spanning multiple imaging modalities with both general and modality-specific degradations at varying severity levels. Extensive experiments demonstrate that PGE-SAM achieves state-of-the-art robustness in both medical and natural image domains across multiple degradation levels, while maintaining generalization to clean images and adding less than one-fifth of the parameters of prior methods.

\end{abstract}

\section{Introduction}
\label{sec:intro}

\begin{figure}[tb]
\centering
\includegraphics[width= 0.9\linewidth]{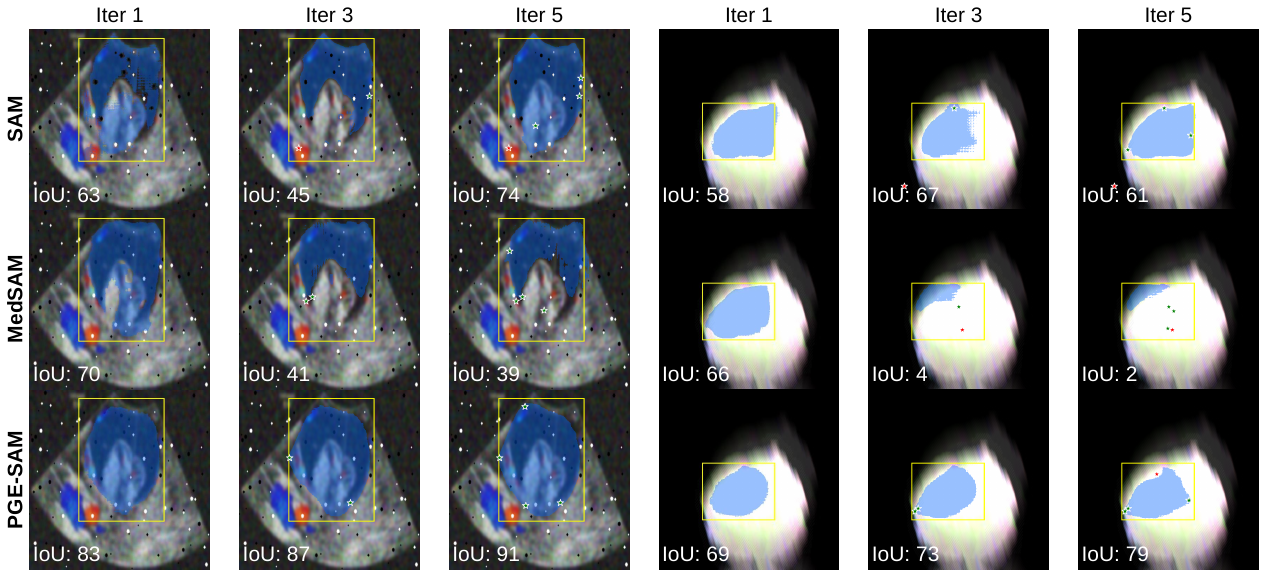}
   \caption{\textbf{Interactive segmentation under degradation.} SAM and MedSAM struggle on a degraded medical image across interaction steps, producing inaccurate masks that fail to improve with refinement. Our \mName{} consistently produces high-quality masks with increasing IoU at each step.}
\label{fig:viz_intro}
\end{figure}

Image degradations such as noise, blur, low resolution, and compression artifacts are ubiquitous in real-world visual data~\cite{hendrycks2019robustness,iayolo,Gupta_2024_WACV,MODE}. These distortions present significant challenges for downstream vision tasks, as models trained on clean data often suffer substantial performance drops when applied to degraded inputs.

The Segment Anything Model (SAM)~\cite{sam} has emerged as a transformative foundation model for promptable image segmentation, demonstrating impressive zero-shot generalization by training on over one billion masks. A key strength of SAM lies in its {interactive, iterative} design: users provide prompts (\eg, clicks or bounding boxes) and progressively refine the prediction through successive interactions, making it particularly suited for annotation and clinical workflows where precision is paramount. However, as illustrated in Fig.~\ref{fig:viz_intro}, SAM's performance degrades markedly under imaging degradations~\cite{sam_robustness1,sam_robustness2}, producing unreliable masks even after multiple refinement steps.

Recent works have attempted to enhance SAM's robustness under degraded inputs. RobustSAM~\cite{robustsam} introduces anti-degradation modules trained with a consistency loss to align features from degraded images with those extracted from clean counterparts. GleSAM~\cite{glesam} adopts a generative strategy by leveraging a pretrained denoising U-Net from Stable Diffusion~\cite{stable_diffusion} to restore corrupted encoder features in a single forward pass. Although these approaches improve segmentation performance under corruption, they suffer from two key limitations. First, feature restoration is performed globally over the entire spatial domain. Background regions that are irrelevant to the segmentation target are reconstructed together with the foreground. This global alignment does not directly support the downstream segmentation objective and may even introduce unwanted artifacts. 
Second, both methods treat enhancement as a one-shot process. They do not account for SAM's iterative refinement mechanism. In practice, a single interaction is often insufficient, especially under severe degradation. Interactive segmentation requires progressive refinement across multiple interactions. Consequently, these methods may deteriorate when additional interactions are provided, despite the expectation of improvement.

To address these limitations, we propose \textbf{P}rompt-\textbf{G}uided Feature \textbf{E}nhancement SAM (\textbf{\mName{}}), a framework that explicitly leverages SAM's interactive and iterative nature to guide the feature restoration process. At its core is the \textbf{Prompt Guidance Generator (PGG)}, which aggregates user prompts (points, boxes) and mask predictions from prior iterations to generate a spatial guidance map that directs the enhancement toward the {region of interest}. This design directly benefits the downstream segmentation task by concentrating the restoration effort on regions that matter, while suppressing background noise. To recover fine-grained details that are often lost in the deeper stages of the backbone, particularly under degradation, we propose \textbf{Multi-Scale Features Interaction (MSFI)}, which integrates low-level features from early encoder layers into the enhancement pipeline. We further introduce a \textbf{Foreground Reconstruction (FR) Loss} that restricts the feature-level supervision to the foreground mask region, replacing the global reconstruction objectives of prior works. Our framework adds less than one-fifth of the parameters introduced by GleSAM and RobustSAM, while achieving substantial performance gains on both medical and natural image benchmarks.

Furthermore, existing degradation-robust segmentation methods have focused exclusively on the natural image domain, overlooking the medical imaging domain where accurate segmentation and model robustness are arguably even more critical. Medical images suffer not only from common degradations such as Gaussian and motion blur, but also from domain-specific artifacts, \eg, photon starvation in low-dose CT and acoustic shadowing in ultrasound images~\cite{photon_starv,acoustic_ultrasound}. To the best of our knowledge, we are the first to investigate interactive segmentation under degradation in the medical imaging domain. To this end, we construct \textbf{\dataName}, an extensive benchmark spanning multiple imaging modalities including CT, MRI, and X-ray, derived from existing medical segmentation datasets. We develop a degradation simulation framework that synthesizes both general imaging artifacts and modality-specific degradations at multiple severity levels. We expect \dataName\ to foster the development of more robust segmentation models and support future research in medical image analysis.

Overall, our contributions are summarized as follows:
\begin{itemize}
    \item We propose {\mName{}, a prompt-guided feature enhancement framework} for SAM that explicitly leverages interactive prompts and iterative mask predictions to spatially direct the restoration process toward regions of interest, thereby directly benefiting the downstream segmentation task.
    \item We construct {the \dataName\ dataset}, comprising diverse medical imaging modalities with multi-level degradations, and develop a degradation simulation framework covering both general and modality-specific artifacts to effectively train and evaluate robust segmentation models.
    \item {Extensive experiments} demonstrate that \mName{} achieves state-of-the-art robustness across both medical and natural image domains under multiple degradation levels, while maintaining strong generalization to clean images and requiring significantly fewer added parameters than prior methods.
\end{itemize}

\section{Related Work}
\label{sec:related_work}

\textbf{Robust Segment Anything.}
The Segment Anything Model (SAM)~\cite{sam} established a new paradigm for promptable image segmentation, enabling strong zero-shot generalization. SAM-HQ~\cite{sam_hq} further improves mask quality via a High-Quality Output Token. However, both assume clean inputs and degrade significantly under noise, blur, or low resolution.
To address this, RobustSAM~\cite{robustsam} introduces a Robust Output Token with adapter layers, trained with a consistency loss that aligns degraded and clean features. GleSAM~\cite{glesam} leverages a pretrained denoising U-Net from Stable Diffusion~\cite{stable_diffusion} to restore degraded encoder features in a single forward pass. As an orthogonal approach, GaRA-SAM~\cite{garasam} introduces gated-rank adaptation, where lightweight adapters dynamically adjust the effective rank of their weight matrices based on the input; however, no pretrained weights have been publicly released at the time of submission.
These feature-alignment methods neglect SAM's iterative refinement and restore features globally rather than focusing on the segmentation target, limiting their practical effectiveness. In contrast, our \mName{} explicitly leverages user prompts and iterative mask predictions to spatially guide restoration, aligning feature enhancement directly with the segmentation objective.

\textbf{SAM for Medical Image Segmentation.}
The zero-shot capability of SAM has spurred numerous adaptations for medical image segmentation~\cite{nnsam,desam,sam_adapter,sam_med3d,medsam2,medsam3,resurgsam2}. MedSAM~\cite{medsam} fine-tunes the full SAM on over 1.5M image-mask pairs across ten modalities, while SAM-Med2D~\cite{sammed2d} scales to 19.7M masks with adapter modules in the encoder for efficient domain adaptation. LiteMedSAM~\cite{medsam} further distills MedSAM into a compact Tiny-ViT backbone for deployment in resource-constrained clinical settings. Despite their success, all existing medical SAM adaptations assume artifact-free inputs and do not account for the degradations prevalent in real-world acquisition, including common artifacts such as motion blur and noise, as well as modality-specific degradations, \eg, photon starvation in low-dose CT and metallic artifacts in X-ray. Our work bridges this gap by equipping SAM with degradation-robust feature enhancement while retaining compatibility with established medical SAM backbones.

\section{Methodology}
\label{sec:method}

\begin{figure}[tb]
\centering
\includegraphics[width=1.0\linewidth]{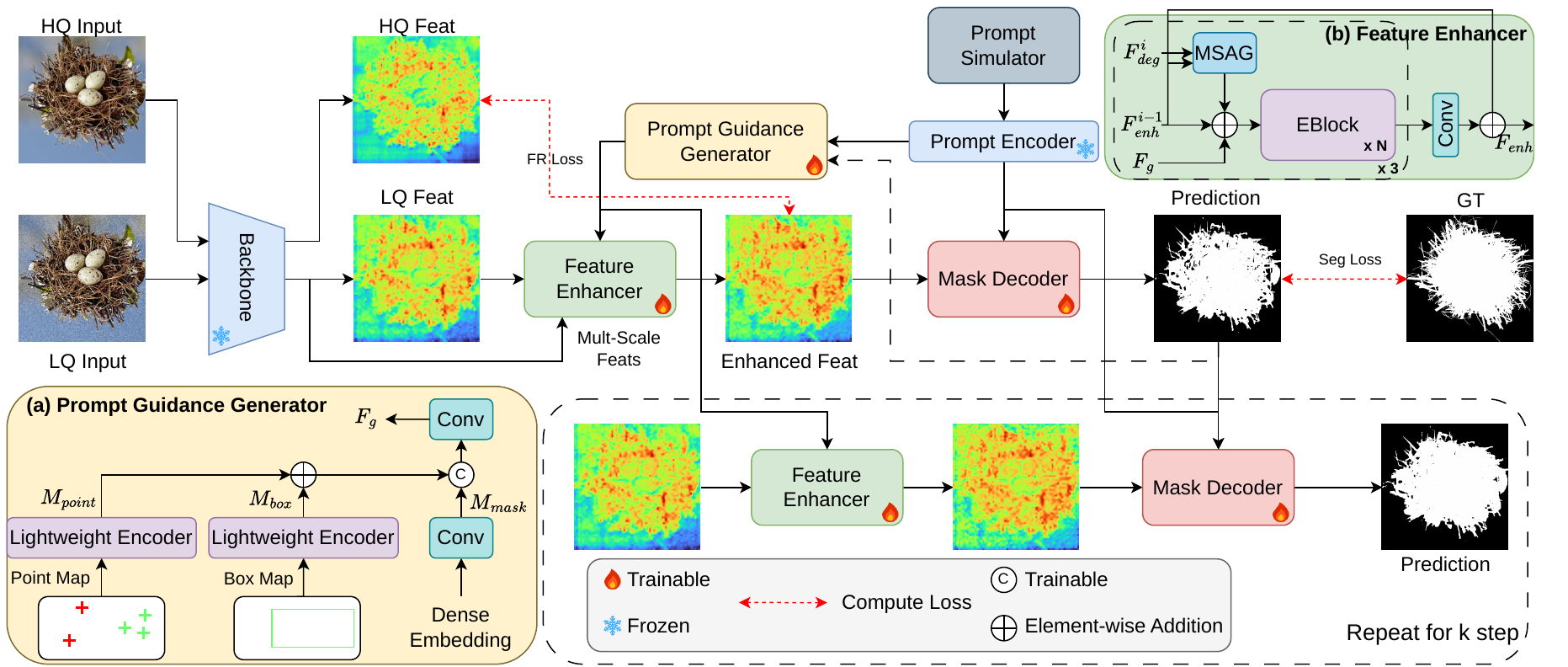}
   \caption{\textbf{Overview of \mName{}.} During training, a paired HQ and degraded image are processed by a shared frozen backbone. The proposed \textbf{Feature Enhancer} restores the degraded feature guided by the \textbf{Prompt Guidance Generator} (PGG), which encodes user prompts and prior mask predictions as spatial guidance $F_g$. The restored feature is decoded by the Mask Decoder and jointly supervised by a segmentation loss and the proposed Foreground Reconstruction (FR) Loss. The process is repeated iteratively for $K$ steps. The HQ branch is discarded at inference. Insets (a) and (b) show the detailed architectures of the PGG and Feature Enhancer, respectively.}
\label{fig:Overview}
\end{figure}

We present \textbf{P}rompt-\textbf{G}uided Feature \textbf{E}nhancement SAM (\mName{}), a novel framework designed to adapt SAM for robust performance under severe imaging degradations. In contrast to existing strategies~\cite{robustsam,glesam} that rely on static, global feature alignment, our approach treats restoration as a dynamic, iterative process aligned with SAM's interactive inference loop. To this end, we introduce a \textbf{Prompt Guidance Generator}, which explicitly leverages user prompts and mask predictions from previous iterations to spatially constrain feature enhancement to the relevant regions of interest. This mechanism effectively suppresses background noise while preserving semantic integrity. Complementing this, we propose a \textbf{Multi-Scale Features Interaction} module to recover fine-grained boundary details often lost in deep semantic layers, and a \textbf{Foreground Reconstruction Loss} to enforce high-fidelity restoration specifically within the target object area. An overview of the proposed architecture is illustrated in Fig.~\ref{fig:Overview}.

In the training phase, given a pair of clear and degraded images, we first extract their feature representations using a frozen SAM Image Encoder. Specifically, we obtain the high-level bottleneck features $F_{\text{clr}}, F_{\text{deg}} \in \mathbb{R}^{C \times H \times W}$ and a set of multi-scale intermediate features $\{F_{\text{deg}}^i\}_{i=0}^2$ from the degraded input. A learnable lightweight Feature Enhancer is then employed to transform the degraded features into a restored representation $F_{\text{enh}} \in \mathbb{R}^{C \times H \times W}$. Crucially, to better align with the downstream segmentation task, we integrate the Prompt Guidance Generator (PGG). The PGG aggregates user prompts (points, boxes) and, during iterative refinement, the mask logits from the preceding step to generate a spatial attention map. This explicitly directs the restoration process toward the semantic regions of interest. Subsequently, the enhanced feature $F_{\text{enh}}$, along with sparse and dense prompt embeddings, is fed into the mask decoder to predict the segmentation mask. The framework is optimized end-to-end: the final mask is supervised via standard segmentation losses, while the enhanced feature $F_{\text{enh}}$ is aligned with $F_{\text{clr}}$ using our proposed Foreground Reconstruction Loss. During inference, the clear image branch is discarded, and the model operates solely on degraded inputs.

\subsection{Prompt Guidance Generator (PGG)}

Rather than restoring features globally, we explicitly guide enhancement using SAM's interactive prompts.
Specifically, we leverage user prompts (points and boxes) with the mask prediction from the previous iteration to construct a spatial guidance map, denoted as $F_g \in \mathbb{R}^{C \times H \times W}$. This guidance map provides strong spatial priors to the restoration network and encourages a focus on the target object rather than irrelevant background regions.

As depicted in Fig.~\ref{fig:Overview}(a), PGG synthesizes sparse geometric cues and dense semantic cues. 
For {point prompts}, we generate a spatial confidence map by modeling positive and negative clicks as Gaussian distributions centered at the click coordinates.
For {box prompts}, we similarly apply a Gaussian kernel at the box center, scaled by the box dimensions, to encode spatial proximity. 
These spatial maps are projected to the feature dimension $C$ via a lightweight encoder to yield $M_{\text{point}}, M_{\text{box}} \in \mathbb{R}^{C \times H \times W}$. To incorporate the {mask prompt}, we extract the dense embeddings from the frozen Prompt Encoder. These are passed through a projection layer to obtain $M_{\text{mask}} \in \mathbb{R}^{C \times H \times W}$.
Finally, we fuse these components to generate the unified guidance feature $F_g$. We first aggregate the geometric priors via element-wise summation, concatenate them with the mask prior, and fuse them using a lightweight convolution layer. This process is formulated as:
\begin{equation}
    F_g = \mathrm{Conv} \left( \mathrm{Concat}\left[ (M_{\text{point}} + M_{\text{box}}), M_{\text{mask}} \right] \right).
\end{equation}
$F_g$ is later injected into the enhancement process via residual addition, effectively highlighting the RoI while suppressing artifacts in the background.

\subsection{Multi-Scale Features Interaction (MSFI)}

Prior methods such as GleSAM~\cite{glesam} rely solely on the final-stage feature for image restoration. However, this deepest-layer representation is single-scale and primarily encodes high-level semantics. It lacks high-frequency details, such as edges and textures, which are crucial for precise segmentation. This issue becomes more severe under degraded conditions, where fine-grained structures are already heavily corrupted.
We posit that both high-level semantic information from deeper layers and low-level spatial details from earlier layers are essential for effective image enhancement.

Specifically, we extract features after the third, second, and first global attention blocks of the ViT encoder, denoted as $\{F_{\text{deg}}^{i}\}_{i=0}^{2}$, and leverage them as intermediate representations to provide complementary information during the enhancement process. The Feature Enhancer is organized into three sequential stages, each comprising $N$ Enhancement Blocks (EBlock) and a Multi-Scale Attention Gate (MSAG). At each stage $i$, the MSAG first modulates the corresponding intermediate feature $F_{\text{deg}}^{i}$ conditioned on the current enhanced representation, thereby selectively controlling the flow of information from earlier layers. The modulated feature is then combined with the current enhanced feature $F_{\text{enh}}^{i-1}$ and the guidance feature $F_{g}$ via element-wise summation, and subsequently refined through the EBlock. Finally, the output of the last stage is projected and combined with the original degraded feature $F_{\text{deg}}$ through a residual connection. Formally, this process is defined as:
\begin{gather}
    F_{\text{enh}}^{i} = \mathrm{EBlock}\!\left(F_{\text{enh}}^{i-1} + F_{g} + \mathrm{MSAG}(F_{\text{deg}}^{i},\; F_{\text{enh}}^{i-1})\right), \quad i = 0, 1, 2 \label{eq:eblock} \\
    F_{\text{enh}} = F_{\text{deg}} + \mathrm{Conv}\!\left(F_{\text{enh}}^{2}\right) \label{eq:residual}
\end{gather}
where $F_{\text{enh}}^{-1} \triangleq F_{\text{deg}}$ serves as the initial input to the first stage. The overall process is illustrated in Fig.~\ref{fig:Overview}(b).

\begin{figure}[tb]
\centering
\includegraphics[width=0.8\linewidth]{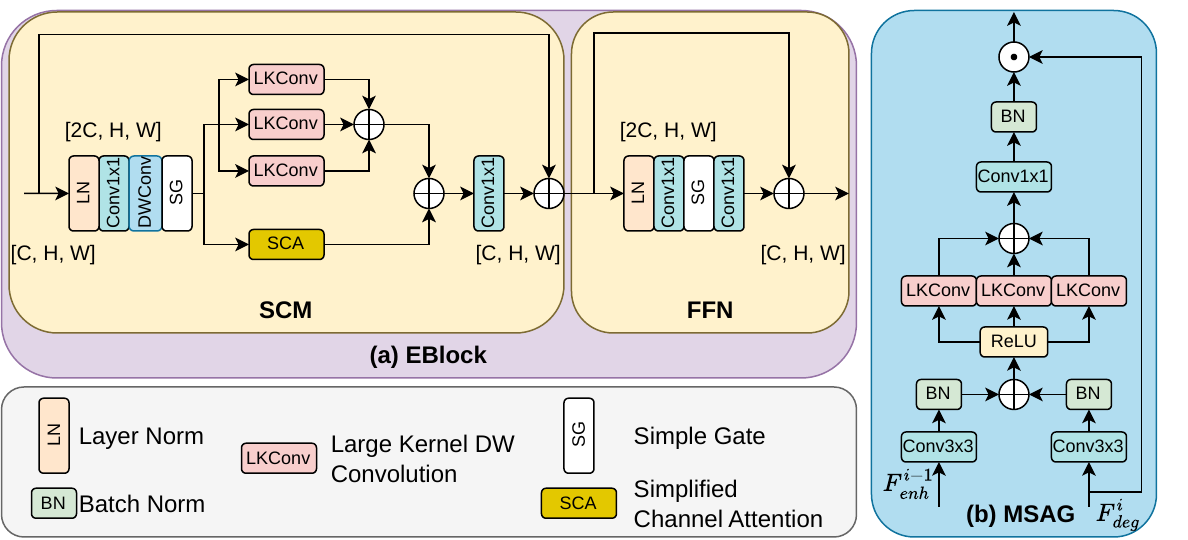}
   \caption{\textbf{Architectures of the Enhancement Block (EBlock) and Multi-Scale Attention Gate (MSAG).} \textbf{(a)} The EBlock consists of a Spatial-Channel Module (SCM) and a Feed-Forward Network (FFN). \textbf{(b)} The MSAG takes the intermediate encoder feature $F_{\text{deg}}^{i}$ and the current enhanced feature $F_{\text{enh}}^{i-1}$ as inputs, and produces a soft attention map to selectively gate the flow of information.} 
\label{fig:EB}
\end{figure}

\textbf{Enhancement Block (EBlock).}
It adopts the overall architecture of NAFBlock~\cite{nafnet}, consisting of a Spatial-Channel Module (SCM) followed by a feed-forward network (FFN), each wrapped in a residual connection:
\begin{gather}
    z_1 = \mathrm{SCM}(\mathrm{LayerNorm}(z)) + z, \label{eq:eblock_scm} \\
    z_2 = \mathrm{FFN}(\mathrm{LayerNorm}(z_1)) + z_1. \label{eq:eblock_ffn}
\end{gather}
The SCM operates along two parallel branches for spatial and channel information aggregation, respectively. In the spatial branch, we employ multiple parallel depth-wise convolutions with varying kernel sizes to capture local patterns across different receptive fields. In the channel branch, we adopt the Simplified Channel Attention mechanism from~\cite{nafnet} to model global inter-channel dependencies. The outputs of both branches are aggregated via element-wise summation and projected through a $1{\times}1$ convolution before being passed to the FFN. The detailed architecture is illustrated in Fig.~\ref{fig:EB}(a).

\textbf{Multi-Scale Attention Gate (MSAG).}
It is inspired by the Attention Gate~\cite{attentiongate} but introduces two key modifications. First, we replace the $1{\times}1$ projection convolutions with $3{\times}3$ convolutions to incorporate local spatial context during gating. Second, we augment the gate with multi-scale parallel large-kernel depth-wise convolutions, enabling the module to capture fine-grained local structural information across multiple receptive fields. Given the intermediate encoder feature $F_{\text{deg}}^{i}$ and the current enhanced feature $F_{\text{enh}}^{i-1}$ as the gating signal, MSAG produces a soft attention map that selectively weights the intermediate feature, retaining only the most informative spatial regions relevant to the current stage of enhancement. This mechanism allows the network to adaptively control the flow of multi-scale information, suppressing redundant or noisy activations from earlier encoder layers. The detailed architecture is depicted in Fig.~\ref{fig:EB}(b).

\subsection{Training Objectives}
\label{sec:training}

We jointly optimize the Feature Enhancer and SAM's Mask Decoder in an end-to-end manner, while keeping the Image Encoder frozen. The total training objective comprises a segmentation loss applied to the predicted masks and a feature-level reconstruction loss that supervises the enhancement module.

\textbf{Foreground Reconstruction Loss.}
Prior works~\cite{robustsam,glesam} apply a global MSE loss over the entire feature map, forcing reconstruction of background regions irrelevant to the segmentation target. We propose the {Foreground Reconstruction} (FR) Loss, which restricts feature-level supervision to the foreground region defined by the ground-truth mask $G$, directly linking restoration to the segmentation objective. Formally:
\begin{equation}
    \mathcal{L}_{\text{FR}}(F_{\text{enh}}, F_{\text{clr}}, G) = \frac{\| (F_{\text{enh}} - F_{\text{clr}}) \odot G \|_2}{\| G \|_1 + \epsilon},
    \label{eq:fr_loss}
\end{equation}
where $\odot$ denotes element-wise multiplication with the broadcasted binary mask $G$, and $\epsilon$ is a small constant, i.e., $10^{-8}$, for numerical stability. By masking out the background, the FR Loss directs the network to prioritize faithful restoration of task-relevant features, effectively aligning the enhancement process with the segmentation objective.

\textbf{Overall Loss.}
The total training loss combines both Dice loss \cite{dice_loss} and Focal loss \cite{focal_loss} for mask supervision with the proposed FR Loss for feature-level supervision:
\begin{equation}
    \mathcal{L}_{\text{total}} = \mathcal{L}_{\text{dice}}(P, G) + \lambda_1 \mathcal{L}_{\text{focal}}(P, G) + \lambda_2 \mathcal{L}_{\text{FR}}(F_{\text{enh}}, F_{\text{clr}}, G),
    \label{eq:total_loss}
\end{equation}
where $P$ denotes the predicted mask, $G$ is the ground-truth mask, and $\lambda_1$, $\lambda_2$ are weighting coefficients that balance the contribution of each loss term.

\section{DM-Seg: A Degraded Multi-Modal Medical Segmentation Benchmark}
\label{sec:deg_med_dataset}
\subsection{Dataset Collection and Organization}

We collected and unified 17 medical image segmentation datasets, which are partitioned into two categories based on their role in evaluation:

\textbf{Seen Datasets.} Twelve datasets are used for training with a 20\% validation split to evaluate in-domain restoration performance. These datasets cover multiple medical imaging modalities, including ultrasound imaging~\cite{vitale2020improving, al2020dataset, pedraza2015open, ultrasound-nerve-segmentation, marzola2021deep, zhao2022mmotu}, brain MRI for tumor analysis~\cite{baid2021rsna}, abdominal CT~\cite{ma2022fast}, dental imaging~\cite{silva2018automatic}, dermoscopic skin lesion images~\cite{codella2018skin}, chest X-ray~\cite{seibold2022detailed}, and microscopy images for cell nuclei segmentation~\cite{caicedo2019nucleus}.

\textbf{Unseen Datasets.} Five datasets are held out from training to evaluate cross-domain generalization. These datasets span diverse medical imaging modalities, including CT/MRI~\cite{kavur2021chaos}, colonoscopy images for polyp segmentation~\cite{jha2019kvasir, bernal2015wm}, and brain MRI for low-grade glioma segmentation~\cite{buda2019association}.

\subsection{Degradation Simulation Framework}

Our pipeline implements two categories of degradations to comprehensively simulate real-world image quality issues. \textbf{General degradations} include eight types commonly observed across all modalities: Gaussian noise, salt \& pepper noise, motion blur, Gaussian blur, brightness/contrast variations, gamma correction, JPEG compression artifacts, and glare. \textbf{Domain-specific degradations} simulate modality-specific artifacts. For example, CT images commonly suffer from photon starvation and beam hardening~\cite{mori2013photon}, while ultrasound images frequently exhibit speckle noise~\cite{al2024accurate}. Complete specifications are provided in the supplementary material.

\subsection{Multi-Level Degradation Simulation}

We generate four levels for each input image with progressively increasing degradation severity: i) \textbf{LQ-1 to LQ-3}: increasing levels of general noise; ii) \textbf{LQ-3+ (Heavy + Domain)}: LQ-3-level degradation with domain-specific artifacts.
For each image, $k \sim \mathcal{U}\{2, 4\}$ degradation types are randomly sampled from the general set and applied in random order, with parameters drawn from level-specific ranges. The same degradation types are applied across LQ-1--LQ-3 for semantic consistency, varying only in intensity. For LQ-3+, domain-specific degradations are subsequently applied based on the imaging modality. For each original image, five versions are generated (one clean and four degraded), resulting in a 5$\times$ expansion of effective training data. Complete specifications and domain-specific simulator implementations are provided in the supplementary document.

\section{Experiment}
\label{sec:expe}

\subsection{Datasets}

\textbf{Medical Image Domain.}
We benchmark \mName{} in the medical image domain using our proposed \textbf{\dataName}\ dataset (Sec.~\ref{sec:deg_med_dataset}), which spans multiple imaging modalities and includes modality-specific degradation patterns. We evaluate on both seen and unseen test splits across four degradation levels (LQ-1 through LQ-3+), covering a wide range of image quality conditions in clinical settings.

\textbf{Natural Image Domain.}
We additionally evaluate \mName{} on natural images, following the data protocol of GleSAM~\cite{glesam}. For training, we use three datasets: LVIS~\cite{lvis}, MSRA-10K~\cite{msra}, and ThinObject-5K~\cite{thin}, comprising approximately 30K image-mask pairs in total, with multi-level degradation synthesized as random combinations of common degradation models. For evaluation, the seen sets consist of the ThinObject-5K and LVIS evaluation splits, while ECSSD~\cite{ecssd} and COCO-val~\cite{coco} serve as unseen sets. All models are evaluated across three degradation levels following the protocol of GleSAM.

\subsection{Experimental Setup}

\textbf{Baselines.}
On DM-Seg, we compare \mName{} against a diverse set of interactive segmentation methods spanning three categories.
\textbf{(1) General-purpose models:} SAM~\cite{sam} and the lightweight Tiny-ViT backbone of MobileSAM~\cite{mobile_sam}.
\textbf{(2) Medical-domain models:} SAM-Med2D~\cite{sammed2d} (with and without adapter layers) and MedSAM~\cite{medsam} along with its lightweight variant, LiteMedSAM, which are SAM-based models fine-tuned on large-scale medical imaging data.
\textbf{(3) Degradation}\textbf{-robust models:} RobustSAM~\cite{robustsam} and GleSAM~\cite{glesam}, which are specifically designed to handle degraded input images.
On LQ-Seg, we compare against SAM, RobustSAM, and GleSAM, as well as cascade baselines that apply state-of-the-art all-in-one image restorers MoCE-IR and DCPT~\cite{moceir,dcpt} as a preprocessing step prior to SAM inference. We note that this cascade paradigm is evaluated only on natural images, as, to the best of our knowledge, no publicly available pretrained weights currently exist for all-in-one medical image restoration.

\textbf{Prompt Simulation.}\label{par:prompt_sim}
Since interactive segmentation models rarely achieve satisfactory results in a single interaction, we simulate five iterative steps per training example, following the protocol of~\cite{scribbleprompt}. At the first step, the prompt type is selected randomly from clicks or a bounding box. {For click prompts}, we initialize with at least one positive click and zero or more negative clicks; each point is sampled uniformly at random from one of three strategies: random spatial sampling, sampling at the centroid of the ground-truth mask, or sampling at the center of an error region from a prior prediction. {For box prompts}, we sample a bounding box around the ground-truth mask with random jitter that can either enlarge or shrink the box relative to the tight ground-truth bounding box. At each subsequent step, a single corrective click (positive or negative) is added based on the current prediction error.

\textbf{Evaluation Strategy and Metrics.}
All models are evaluated using the prompt simulation protocol described above. For MedSAM and LiteMedSAM, we additionally report a box-prompt-only variant, as these models exhibit substantially stronger performance under this prompt type. We report four complementary metrics: IoU, Dice, Pixel Accuracy and Normalized Surface Dice (NSD).

\textbf{Implementation Details.}
For all baselines, images are resized to each model's required input resolution, and predictions are upsampled back to $1024 \times 1024$ for evaluation against ground-truth masks. Our models are trained for 36 epochs on $4{\times}$ A100 GPUs using the AdamW optimizer with a base learning rate of $5 \times 10^{-4}$. The loss weighting coefficients are empirically set to $\lambda_1 = 20$ and $\lambda_2 = 5$ in Eq.~\ref{eq:total_loss}. For medical images, \mName{} and Lite\mName{} use pretrained MedSAM and LiteMedSAM backbones, respectively; for natural images, \mName{} uses the pretrained SAM ViT-B backbone. Training details for each model variant are provided in the supplementary material.

\subsection{Main Results}

\begin{table}[tb]
\centering
\caption{\textbf{Comparison on the seen split of \dataName} across five degradation levels (LQ-3+ to Clear). All methods are evaluated with five iterative prompt steps using IoU and Dice (\%). * denotes no adapter version, $^\dagger$ denotes box-only variant. \textbf{Bold} and \underline{underline} indicate the best and second-best results, respectively.}
\setlength{\tabcolsep}{3.5pt}
\tiny
\begin{tabular}{ccccccccccccc}\toprule[1.pt]
     \multirow{2}{*}{{Method}} & \multicolumn{2}{c}{{LQ-3+}} & \multicolumn{2}{c}{{LQ-3}} &\multicolumn{2}{c}{{LQ-2}} & \multicolumn{2}{c}{{LQ-1}} & \multicolumn{2}{c}{{Clear}} &\multicolumn{2}{c}{{Average}} \\
\cmidrule(r){2-3}  \cmidrule(r){4-5} \cmidrule(r){6-7} \cmidrule(r){8-9}  \cmidrule(r){10-11} \cmidrule(r){12-13}
                               & IoU & Dice & IoU & Dice 
                            & IoU & Dice & IoU & Dice & IoU & Dice 
                            & IoU & Dice \\
\toprule[1.pt]
\multicolumn{13}{c}{{Natural Image Domain}} \\
\midrule
MobileSAM   & 60.92 & 71.35 & 63.03 & 72.93 & 67.11 & 76.34 & 69.09 & 77.85 & 69.36 & 78.01 & 65.90 & 75.30 \\
SAM & 63.14 & 73.33 & 65.24 & 74.90 & 68.21 & 77.20 & 69.77 & 78.42 & 70.00 & 78.45 & 67.27 & 76.46 \\
\midrule
\multicolumn{13}{c}{{Medical Image Domain}} \\
\midrule
SAM-Med2D & 41.46 & 52.69 & 43.25 & 53.93 & 47.22 & 57.25 & 49.09 & 58.98 & 48.95 & 58.40 & 45.99 & 56.25 \\
SAM-Med2D*    & 60.03 & 70.52 & 62.20 & 72.28 & 63.42 & 73.19 & 61.55 & 71.48 & 60.49 & 70.48 & 61.54 & 71.59 \\
LiteMedSAM       & 24.68 & 34.34 & 25.38 & 34.75 & 29.48 & 39.40 & 33.48 & 43.82 & 34.48 & 44.54 & 29.50 & 39.37 \\
MedSAM           & 37.12 & 48.67 & 37.94 & 49.37 & 39.63 & 50.79 & 42.11 & 53.22 & 44.43 & 55.34 & 40.24 & 51.48 \\
LiteMedSAM$^\dagger$       & 53.71 & 64.84 & 54.15 & 64.94 & 57.59 & 67.88 & 59.42 & 69.16 & 62.48 & 71.25 & 57.47 & 67.61 \\
MedSAM$^\dagger$           & 65.58 & 75.40 & 67.41 & 76.77 & 69.63 & 78.39 & 72.21 & 80.27 & 74.91 & 82.08 & 69.95 & 78.58 \\
\midrule
\multicolumn{13}{c}{{Degraded Image Domain}} \\
\midrule
RobustSAM        & 25.50 & 35.25 & 26.23 & 35.84 & 26.75 & 36.35 & 26.67 & 36.26 & 27.19 & 36.87 & 26.47 & 36.11 \\
GleSAM           & 15.34 & 23.57 & 15.97 & 24.25 & 16.52 & 24.89 & 16.60 & 25.00 & 17.20 & 25.76 & 16.33 & 24.69 \\
\rowcolor{lightgray!40}
Lite\mName{}    & \underline{77.22} & \underline{85.05} & \underline{77.95} & \underline{85.50} & \underline{79.71} & \underline{86.57} & \underline{80.51} & \underline{87.10} & \underline{80.61} & \underline{87.14} & \underline{79.20} & \underline{86.27} \\
\rowcolor{lightgray!40}
\mName{} & \textbf{82.22} & \textbf{89.18} & \textbf{82.85} & \textbf{89.51} & \textbf{84.65} & \textbf{90.63} & \textbf{85.31} & \textbf{91.03} & \textbf{85.82} & \textbf{91.32} & \textbf{84.17} & \textbf{90.33} \\
\toprule[1.pt]
\end{tabular}
\label{tab:seen_dataset}
\end{table}

\begin{figure}[t!]
\centering
\includegraphics[width=1.0\linewidth]{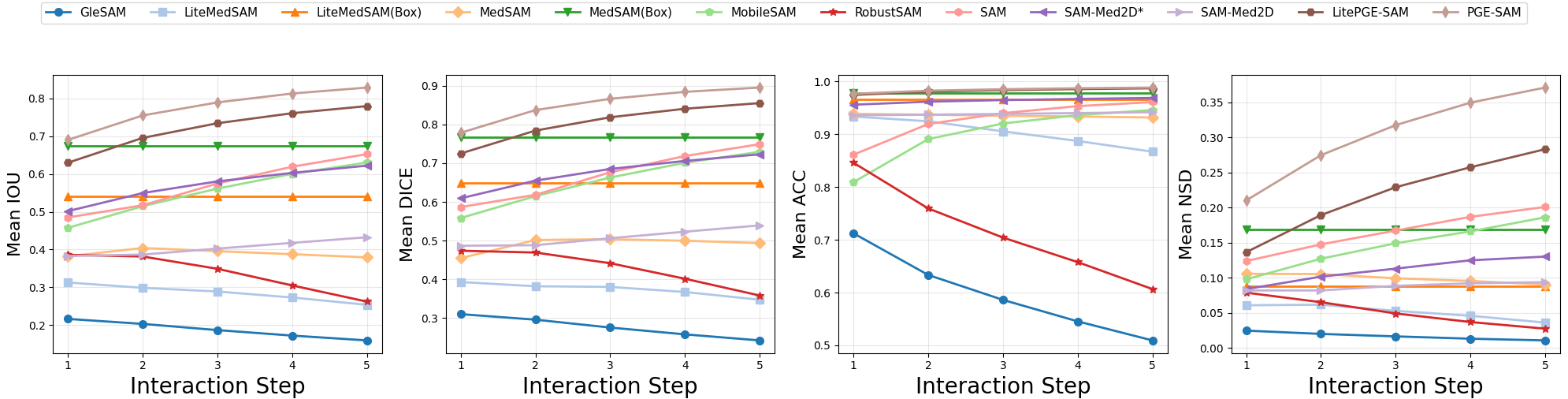}
   \caption{\textbf{Performance across interaction steps.} Mean IoU, Dice, Accuracy, and NSD at each of the five interaction steps under LQ-3 degradation.} 
\label{fig:iterative}
\end{figure}

\begin{table}[tb]
\centering
\caption{\textbf{Performance comparison on the unseen split of \dataName} across five degradation levels (LQ-3+ to Clear). All methods are evaluated with five iterative prompt steps using IoU and Dice (\%). * denotes no adapter, $^\dagger$ denotes box-only variant. \textbf{Bold} and \underline{underline} indicate the best and second-best results, respectively.}
\setlength{\tabcolsep}{3.5pt}
\tiny
\begin{tabular}{ccccccccccccc}\toprule[1.pt]
     \multirow{2}{*}{{Method}} & \multicolumn{2}{c}{{LQ-3+}} & \multicolumn{2}{c}{{LQ-3}} &\multicolumn{2}{c}{{LQ-2}} & \multicolumn{2}{c}{{LQ-1}} & \multicolumn{2}{c}{{Clear}} &\multicolumn{2}{c}{{Average}} \\
\cmidrule(r){2-3}  \cmidrule(r){4-5} \cmidrule(r){6-7} \cmidrule(r){8-9}  \cmidrule(r){10-11} \cmidrule(r){12-13}
                               & IoU & Dice & IoU & Dice 
                            & IoU & Dice & IoU & Dice & IoU & Dice 
                            & IoU & Dice \\
\toprule[1.pt]
\multicolumn{13}{c}{{Natural Image Domain}} \\
\midrule
MobileSAM   & 58.71 & 69.75 & 61.17 & 71.70 & 67.11 & 76.64 & 71.83 & 80.39 & 77.55 & 84.77 & 67.27 & 76.65 \\
SAM         & 60.91 & 71.78 & 63.02 & 73.41 & \underline{69.28} & \underline{78.49} & \underline{73.89} & \underline{82.14} & \underline{77.95} & \underline{85.10} & \underline{69.01} & \underline{78.18} \\
\midrule
\multicolumn{13}{c}{{Medical Image Domain}} \\
\midrule
SAM-Med2D        & 33.15 & 44.30 & 36.53 & 47.41 & 42.88 & 53.15 & 49.42 & 59.17 & 58.21 & 67.22 & 44.04 & 54.25 \\
SAM-Med2D*       & 49.20 & 60.05 & 52.89 & 63.27 & 56.88 & 66.40 & 59.75 & 68.69 & 62.42 & 70.67 & 56.23 & 65.82 \\
LiteMedSAM       & 22.59 & 31.93 & 22.79 & 31.95 & 27.33 & 36.76 & 33.08 & 42.79 & 40.88 & 50.21 & 29.33 & 38.73 \\
MedSAM           & 33.44 & 45.16 & 34.49 & 46.11 & 37.82 & 49.25 & 42.01 & 53.25 & 46.31 & 57.29 & 38.81 & 50.21 \\
LiteMedSAM$^\dagger$  & 48.25 & 59.87 & 49.14 & 60.27 & 53.71 & 64.17 & 58.05 & 67.79 & 63.69 & 72.02 & 54.57 & 64.82 \\
MedSAM$^\dagger$      & 60.95 & 72.32 & 63.92 & 74.54 & 67.88 & 77.45 & 72.43 & 81.08 & 76.15 & 83.76 & 68.27 & 77.83 \\
\midrule
\multicolumn{13}{c}{{Degraded Image Domain}} \\
\midrule
RobustSAM        & 28.05 & 39.17 & 18.89 & 27.20 & 19.59 & 27.86 & 20.29 & 28.68 & 21.99 & 30.27 & 21.76 & 30.64 \\
GleSAM           & 9.73 & 15.81 & 9.64 & 15.68 & 9.83 & 15.89 & 10.07 & 16.16 & 10.67 & 16.69 & 9.99 & 16.05 \\
\rowcolor{lightgray!40}
Lite\mName       
& \underline{63.41} & \underline{74.80}
& \underline{64.14} & \underline{75.56}
& {65.68} & {76.56}
& {66.57} & {76.90}
& {69.07} & {78.57}
& {65.77} & {76.48} \\

\rowcolor{lightgray!40}
\textbf{\mName}  
& \textbf{71.64} & \textbf{82.18}
& \textbf{72.09} & \textbf{82.46}
& \textbf{73.77} & \textbf{83.53}
& \textbf{75.74} & \textbf{84.77}
& \textbf{78.39} & \textbf{86.57}
& \textbf{74.33} & \textbf{83.90} \\
\toprule[1.pt]
\end{tabular}
\label{tab:unseen_dataset}
\end{table}

\textbf{Performance on Seen \dataName.}
Tab.~\ref{tab:seen_dataset} shows results on the seen split of \dataName. \mName{} achieves the best performance across all degradation levels, with an average Dice of 90.33\% and surpassing the strongest baseline (MedSAM$^\dagger$) by over 11 points. Its lightweight variant, Lite\mName{}, also outperforms all baselines (86.27\% Dice). \mName{} is robust to degradation, with only a 2-point Dice drop from Clear to LQ-3+, compared to 5 for SAM and 7 for MedSAM$^\dagger$.
GleSAM and RobustSAM perform poorly on medical data, which we attribute to both a domain gap and their lack of iterative refinement support. MedSAM and LiteMedSAM, while fine-tuned on medical data, similarly neglect iterative refinement during training, leading to degraded performance under the multi-step evaluation protocol.
We analyze per-step performance in Fig.~\ref{fig:iterative}. \mName{} and Lite\mName{} achieve the highest metrics at every step and consistently improve, confirming our iterative-aware training effectively leverages multi-step refinement. In contrast, GleSAM and LiteMedSAM show declining performance with additional interactions, indicating their single-pass designs are incompatible with iterative prompting.

\begin{table*}[tb]
\centering
\caption{\textbf{Performance comparison on the natural image domain} across three degradation levels on seen (ThinObject-5K, LVIS) and unseen (ECSSD, COCO) datasets with 5 interaction steps.}
\label{tab:natural_image}
\tiny
\setlength{\tabcolsep}{3.5pt}
\begin{tabular}{c|cccccc|cccccc}\toprule[1.pt]
\multirow{3}{*}{Method} & \multicolumn{6}{c|}{ThinObject-5K (Seen)} & \multicolumn{6}{c}{LVIS (Seen)} \\
\cmidrule(r){2-7} \cmidrule(r){8-13}
 & \multicolumn{2}{c}{LQ-3} & \multicolumn{2}{c}{LQ-2} & \multicolumn{2}{c|}{LQ-1} & \multicolumn{2}{c}{LQ-3} & \multicolumn{2}{c}{LQ-2} & \multicolumn{2}{c}{LQ-1} \\
\cmidrule(r){2-3} \cmidrule(r){4-5} \cmidrule(r){6-7} \cmidrule(r){8-9} \cmidrule(r){10-11} \cmidrule(r){12-13}
 & IoU & Dice & IoU & Dice & IoU & Dice & IoU & Dice & IoU & Dice & IoU & Dice \\
\midrule
SAM         & 65.72 & 75.35 & 72.52 & 80.89 & \underline{74.78} & \underline{83.08} & 65.61 & 76.62 & 68.67 & 79.00 & 71.60 & 81.35 \\
MoCE-SAM    & \underline{67.83} & \underline{77.27} & \underline{72.82} & \underline{81.50} & 74.70 & 82.93 & \underline{66.91} & \underline{77.60} & \underline{70.26} & \underline{80.23} & \underline{71.97} & \underline{81.53} \\
DCPT-SAM    & 67.01 & 76.46 & 71.65 & 80.31 & 73.14 & 81.57 & 64.69 & 75.94 & 69.42 & 79.68 & 71.66 & 81.40 \\
RobustSAM   & 42.15 & 55.45 & 44.97 & 57.91 & 45.36 & 58.25 & 26.11 & 35.92 & 26.71 & 36.17 & 30.68 & 40.58 \\
GleSAM      & 48.80 & 60.99 & 50.50 & 62.65 & 52.21 & 64.07 & 18.29 & 26.14 & 19.24 & 27.03 & 21.11 & 29.09 \\
\rowcolor{lightgray!40}
\mName      & \textbf{81.27} & \textbf{88.09} & \textbf{84.29} & \textbf{90.30} & \textbf{85.22} & \textbf{90.91} & \textbf{73.30} & \textbf{83.46} & \textbf{75.49} & \textbf{84.95} & \textbf{76.72} & \textbf{85.84} \\
\toprule[1.pt]
 & \multicolumn{6}{c|}{ECSSD (Unseen)} & \multicolumn{6}{c}{COCO (Unseen)} \\
\midrule
SAM         & 70.26 & 80.59 & 76.26 & 84.95 & 79.36 & 87.08 & 66.23 & 77.26 & \underline{69.62} & \underline{79.81} & \underline{71.97} & \underline{81.52} \\
MoCE-SAM    & \underline{73.54} & \underline{82.97} & \underline{78.94} & \underline{86.77} & \underline{80.75} & \underline{88.05} & \underline{67.07} & \underline{77.85} & 69.57 & 79.66 & 71.04 & 80.80 \\
DCPT-SAM    & 70.38 & 80.51 & 74.77 & 83.82 & 79.77 & 87.39 & 65.21 & 76.43 & 69.53 & 79.73 & 71.23 & 81.02 \\
RobustSAM   & 38.02 & 51.97 & 44.00 & 57.89 & 44.28 & 57.85 & 27.22 & 37.27 & 29.76 & 39.87 & 31.29 & 41.54 \\
GleSAM      & 40.45 & 54.35 & 44.52 & 57.97 & 43.94 & 57.15 & 17.94 & 25.61 & 20.46 & 28.54 & 21.71 & 29.90 \\
\rowcolor{lightgray!40}
\mName      & \textbf{79.91} & \textbf{87.89} & \textbf{83.53} & \textbf{90.27} & \textbf{86.15} & \textbf{92.01} & \textbf{73.56} & \textbf{83.64} & \textbf{75.62} & \textbf{85.08} & \textbf{76.81} & \textbf{85.87} \\
\bottomrule[1.pt]
\end{tabular}
\end{table*}

\textbf{Performance on Unseen \dataName.}
Tab.~\ref{tab:unseen_dataset} evaluates cross-domain generalization on the unseen split of \dataName, comprising datasets not seen during training. \mName{} achieves the highest average Dice of 83.90\%, outperforming the strongest baseline SAM by nearly 6 points. Despite a moderate generalization gap relative to the seen split, \mName{} still surpasses all baselines by a significant margin, confirming that its prompt-guided enhancement generalizes effectively to novel imaging sources.

\textbf{Performance on Natural Image Domain.}
As shown in Tab.~\ref{tab:natural_image}, \mName{} consistently outperforms all baselines across every degradation level on both seen and unseen datasets. RobustSAM and GleSAM, which neglect the iterative nature of SAM during fine-tuning, suffer substantial performance degradation. Notably, the cascade baselines, e.g., MoCE-SAM, DCPT-SAM, offer modest gains under severe degradation but provide diminishing or even negative improvement at milder levels, suggesting that the restoration step introduces artifacts that hinder downstream segmentation.

\subsection{Ablation Studies}
Unless otherwise specified, all ablation experiments are conducted under the LQ-3 degradation level and report IoU and Dice after 5 interaction steps as a representative setting. Due to space constraints, additional ablations are provided in the supplementary material. 

\begin{table}[tb]
  \centering
  \begin{minipage}[t]{0.48\textwidth}
    \centering
    \caption{\textbf{Impact of Components}}
    \label{tab:ablation_components}
    \tiny
    \setlength{\tabcolsep}{4.0pt}
    \renewcommand{\arraystretch}{1.2} 
    \begin{tabular}{ccccc|cc}\toprule[1.pt]
Baseline & FE & PGG & $\mathcal{L}_{\text{FR}}$ & MSFI & IoU & Dice \\
\midrule
\checkmark & & & & & 25.38 & 34.75 \\
\checkmark & \checkmark & & & & 72.22 & 81.41 \\
\checkmark & \checkmark & \checkmark & & & 74.14 & 82.97 \\
\checkmark & \checkmark & \checkmark & \checkmark & & 74.83 & 83.67  \\
\rowcolor{lightgray!40}
\checkmark & \checkmark & \checkmark & \checkmark & \checkmark & \textbf{77.95} & \textbf{85.50} \\
\bottomrule[1.pt]
\end{tabular}
    \label{tab:abs_keyComp}
  \end{minipage}
  \hfill
  \begin{minipage}[t]{0.48\textwidth}
    \centering
    \caption{\textbf{Effect of Architectural Design.}}
    \label{tab:abs_arch}
    \tiny
    \setlength{\tabcolsep}{3.0pt}
    \renewcommand{\arraystretch}{1.2}
    \begin{tabular}{c|cc|c|cc}\toprule[1.pt]
    Method & IoU & Dice & Method & IoU & Dice \\
    \midrule
    \multicolumn{3}{c|}{Enhancement Block} &  \multicolumn{3}{c}{Attention Gate} \\
    \midrule
    NAFBlock & 76.79 & 84.46 &  AG & 76.38 & 84.02 \\
    \rowcolor{lightgray!40}
    EBlock & \textbf{77.95} & \textbf{85.50} & MSAG & \textbf{77.95} & \textbf{85.50} \\
    \bottomrule[1.pt]
    \end{tabular}
  \end{minipage}
\end{table}

\begin{table}[tb]
  \centering
  \begin{minipage}[t]{0.33\textwidth}
    \centering
    \caption{\textbf{Effect of fine-tuning SAM}. The symbol $^\ddagger$ denotes not using seg loss.}
    \label{tab:ablation_ft}
    \tiny
    \setlength{\tabcolsep}{2.0pt}
    \begin{tabular}{l|cc}\toprule[1.pt]
    Method & IoU & Dice \\
    \midrule
    LiteMedSAM & 25.38 & 34.75 \\
    LiteMedSAM + FT Token & 59.00 & 70.69 \\
    LiteMedSAM + FT Mask Dec. & 72.11 & 81.23 \\
    \midrule
    Lite\mName{} + FT FE & 75.66 & 83.82 \\
    Lite\mName{} + FT FE$^\ddagger$ & 18.96 & 27.55 \\
    \rowcolor{lightgray!40}
    Lite\mName{} & \textbf{77.95} & \textbf{85.50} \\
    \bottomrule[1.pt]
    \end{tabular}
  \end{minipage}
  \hfill
  \begin{minipage}[t]{0.65\textwidth}
    \centering
    \caption{\textbf{Efficiency Analysis.} Comparison with other baselines in computational efficiency.}
    \label{tab:efficiency}
    \tiny
    \setlength{\tabcolsep}{4.0pt}
    \begin{tabular}{c|cccc}\toprule[1.pt]
    Method & \begin{tabular}{@{}c@{}}Added \\ Params (M)\end{tabular} & \begin{tabular}{@{}c@{}}Train \\ GPUs\end{tabular} & \begin{tabular}{@{}c@{}}Inf. Mem \\ (MB)\end{tabular} & \begin{tabular}{@{}c@{}}FLOPs \\ (G)\end{tabular} \\
    \midrule
    LiteMedSAM & 0 & - & 873 & 76.81 \\
    LiteRobustSAM & 8.87 & 4 & 975 & 139.51 \\
    \rowcolor{lightgray!40}
    Lite\mName{} & \textbf{4.05} & \textbf{4} & \textbf{903} & \textbf{107.65} \\
    \midrule
    SAM & 0 & 256 & 3566 & 743.98 \\
    RobustSAM & 59.44 & 8 & 3859 & 1012.06 \\
    GleSAM & 2671.56 & 4 & 15338 & 1491.13 \\
    \rowcolor{lightgray!40}
    \mName{} & \textbf{11.75} & \textbf{4} & \textbf{3675} & \textbf{831.06} \\
    \bottomrule[1.pt]
    \end{tabular}
  \end{minipage}
\end{table}

\textbf{Impact of Key Components.}
As shown in Tab.~\ref{tab:abs_keyComp}, each proposed component contributes meaningfully to overall performance. Starting from LiteMedSAM as the baseline, adding and fine-tuning only the Feature Enhancer increases IoU to 72.22, while incorporating the Prompt Guidance Generator further improves IoU to 74.14. This confirms the importance of spatially guided feature enhancement. Replacing the standard reconstruction loss with the proposed $\mathcal{L}_{\text{FR}}$ further improves IoU to 74.83 by focusing restoration on task-relevant regions. Finally, adding the Multi-Scale Feature Interaction module brings the full model to 77.95 IoU, demonstrating that all components are complementary.

\textbf{Effect of Architectural Design.}
To validate the structural choices in our Feature Enhancer, we ablate its core building blocks in Tab.~\ref{tab:abs_arch}. Substituting our Enhancement Block (EBlock) with the baseline NAFBlock~\cite{nafnet} results in a performance drop of over 1.0 point. Furthermore, replacing our MSAG with a standard AG~\cite{attentiongate} leads to a 1.5-point decrease. These results clearly justify our architectural design choices.

\textbf{Effect of Fine-Tuning SAM.}
Tab.~\ref{tab:ablation_ft} compares fine-tuning strategies. Fine-tuning the output token (59.00 IoU) or the full Mask Decoder (72.11 IoU) remains approximately 5 points below \mName{}, indicating that decoder-only adaptation is insufficient under severe degradation. Training the Feature Enhancer without segmentation loss causes a sharp drop to 18.96 IoU, confirming that joint optimization of both modules is essential.

\textbf{Efficiency Analysis.}
Tab.~\ref{tab:efficiency} compares computational efficiency on a single A100 GPU. Lite\mName{} adds only 4M parameters, less than half of LiteRobustSAM (8.87M; a full comparison is provided in the supplementary material), while \mName{} requires 11M added parameters, roughly $5\times$ fewer than RobustSAM and over $200\times$ fewer than GleSAM. Our models also consume less GPU memory and fewer FLOPs than all degradation-specific baselines, achieving state-of-the-art robustness with highly efficient computational profiles.

\subsection{Qualitative Analysis}

\begin{figure}[tb]
\centering
\includegraphics[width=1.0\linewidth]{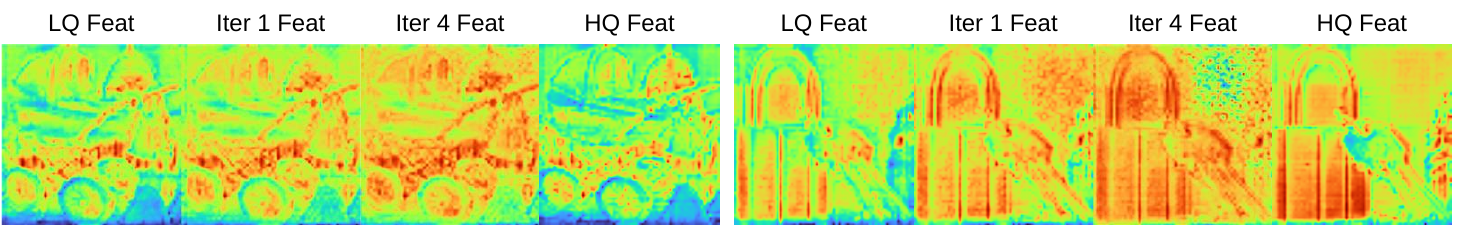}
   \caption{\textbf{Feature visualization across iterative enhancement.} \mName{} progressively sharpens the representation toward the target object with each iteration.} 
\label{fig:feature_viz}
\end{figure}

Fig.~\ref{fig:viz_intro} presents qualitative segmentation comparisons, while Fig.~\ref{fig:feature_viz} visualizes the internal feature representations across iterative enhancement stages. 
As shown in Fig.~\ref{fig:feature_viz}, the enhanced features progressively become more object-focused at each interaction step, demonstrating that prompt guidance effectively steers the restoration toward the region of interest. We note that the enhanced features do not visually replicate the clean reference, as the Foreground Reconstruction Loss serves as an auxiliary objective optimized jointly with the segmentation losses; we posit that the most perceptually faithful reconstruction is not necessarily optimal for the downstream segmentation task. Additional qualitative and quantitative comparisons, per-dataset breakdowns, and implementation details are provided in the supplementary material.

\section{Conclusion}
\label{sec:conclusion}

This paper presents \mName{}, a prompt-aware feature enhancement framework for degradation-robust segmentation. Through prompt-guided spatial enhancement (PGG), multi-scale feature integration (MSFI), and foreground-focused supervision (FR Loss), our approach directs restoration toward task-relevant regions, directly benefiting downstream segmentation. We also introduce \dataName, a benchmark for degraded medical image segmentation spanning multiple modalities with general and modality-specific degradations. Experiments show that \mName{} achieves state-of-the-art performance on both medical and natural images across degradation levels, while adding less than one-fifth of the parameters of prior methods.

\clearpage
\appendix

\title{Supplementary Material: PGE-SAM: Prompt-Guided Feature Enhancement for Interactive Segmentation under Degradation}
\titlerunning{Appendix}
\author{Tuan-Duc Nguyen\inst{1} \and
Anh-Tuan Mai\inst{1} \and
Duc-Trong Le\inst{2}}
\authorrunning{T.~Nguyen et al.}
\institute{FPT Software AI Center \and
VNU University of Engineering and Technology \\
\email{trongld@vnu.edu.vn}}

\maketitle

This supplementary material provides comprehensive supporting analyses and additional results for the main paper. We begin with \textbf{additional ablation studies} (Sec.~\ref{supp:sec:ablations}), which include analyses omitted from the main paper due to space. We also ablate PGG input encoding, the FR loss weight, and the number of EBlocks per stage.

In Sec.~\ref{supp:sec:expe}, we present \textbf{additional experiments} covering performance on clear images, alternative synthetic degradation (RobustSeg-style), zero-shot real-world degradation (LIS, WXSDO, SMDD), backbone scaling (Tiny-ViT to ViT-L), extended 20-step iterative evaluation, sensitivity to initial prompt type, and cross-domain generalization from natural to medical images. Sec.~\ref{supp:sec:imple_detail} details the \textbf{training configurations} for each model variant across both medical and natural image domains. Sec.~\ref{supp:sec:dataset_processing} describes the construction of our \textbf{\dataName{} benchmark}, including dataset composition, modality-specific degradation design, and augmentation parameters. Sec.~\ref{supp:sec:qualitative} provides \textbf{qualitative visualizations} of segmentation results under various degradation conditions. Finally, Sec.~\ref{supp:sec:detailed_iterative} presents \textbf{per-step iterative curves} for every dataset and degradation level, demonstrating that \mName{} outperforms baselines not only through iterative refinement but also at the first interaction step, confirming that its gains are architectural rather than solely a product of iterative training.

\section{Additional Ablation Studies}
\label{supp:sec:ablations}

We present additional ablation studies that complement the analysis in the main paper.

\subsection{Impact of PGG on Full Model}

\begin{table}[tb]
\centering
\caption{\textbf{Impact of PGG on Lite\mName{} (full model) evaluated on \dataName{} (Seen).} Both variants are trained under identical protocols. \textbf{Bold} denotes the best results.}
\setlength{\tabcolsep}{3.0pt}
\tiny
\renewcommand{\arraystretch}{1.2}
\begin{tabular}{ccccccccccccc}\toprule[1.pt]
     \multirow{2}{*}{{Method}} & \multicolumn{2}{c}{{LQ-3+}} & \multicolumn{2}{c}{{LQ-3}} &\multicolumn{2}{c}{{LQ-2}} & \multicolumn{2}{c}{{LQ-1}} & \multicolumn{2}{c}{{Clear}} &\multicolumn{2}{c}{{Average}} \\
\cmidrule(r){2-3}  \cmidrule(r){4-5} \cmidrule(r){6-7} \cmidrule(r){8-9}  \cmidrule(r){10-11} \cmidrule(r){12-13}
                               & IoU & Dice & IoU & Dice 
                            & IoU & Dice & IoU & Dice & IoU & Dice 
                            & IoU & Dice \\
\midrule
Without PGG        & 73.29  & 82.00 & 73.73 & 82.16 & 75.52 & 83.41 & 76.16 & 83.78 & 75.78 & 83.43 & 74.90 & 82.96 \\
\rowcolor{lightgray!40}
With PGG           & \textbf{77.22} & \textbf{85.05} & \textbf{77.95} & \textbf{85.50} & \textbf{79.71} & \textbf{86.57} & \textbf{80.51} & \textbf{87.10} & \textbf{80.61} & \textbf{87.14} & \textbf{79.20} & \textbf{86.27} \\
\bottomrule[1.pt]
\end{tabular}
\label{tab:pgg_abs}
\end{table}

To further isolate the contribution of the Prompt Guidance Generator (PGG), we train the full Lite\mName{} architecture with and without PGG-generated guidance features under identical settings. As reported in Tab.~\ref{tab:pgg_abs}, removing PGG leads to a consistent performance drop across all degradation levels, with an average decrease of 4.30 IoU and 3.31 Dice (e.g., 73.73 vs.\ 77.95 IoU under LQ-3). The improvement from PGG remains substantial across the entire quality spectrum. These results confirm that PGG plays a central role in directing the Feature Enhancer toward task-relevant regions, benefiting segmentation quality regardless of input degradation level.

\subsection{Ablation on Prompt Guidance Generator Inputs}

\begin{table*}[tb]
\centering
\caption{\textbf{Ablation on Prompt Guidance Generator inputs.} We ablate the guidance sources and prompt encoding strategies used in PGG. ``Gaussian'' denotes soft proximity encoding via a Gaussian kernel; ``Binary'' denotes a binary indicator map. \textbf{Bold} denotes the best results.}
\label{tab:pgg_prompt}
\scriptsize
\setlength{\tabcolsep}{1.5pt}
\renewcommand{\arraystretch}{1.2}
\begin{tabular}{ccc|ccccc|ccccc}\toprule[1.pt]
\multirow{2}{*}{Mask} & \multirow{2}{*}{Point Enc.} & \multirow{2}{*}{Box Enc.} & \multicolumn{5}{c|}{LQ-3} & \multicolumn{5}{c}{Average} \\
\cmidrule(r){4-8} \cmidrule(r){9-13}
 & & & IoU$\uparrow$ & Dice$\uparrow$ & PA$\uparrow$ & NSD$\uparrow$ & HD95$\downarrow$ & IoU$\uparrow$ & Dice$\uparrow$ & PA$\uparrow$ & NSD$\uparrow$ & HD95$\downarrow$ \\
\midrule
--- & --- & ---      & 73.73 & 82.16 & 98.10 & 22.41 & 66.64 & 74.90 & 82.96 & 98.21 & 24.26 & 65.69 \\
\checkmark & --- & ---      & 74.61 & 82.78 & 98.40 & 24.66 & 65.56 & 75.96 & 83.66 & 98.51 & 26.80 & 62.71 \\
\checkmark & Binary & Binary     & 76.12 & 84.09 & 98.57 & 25.23 & 58.03 & 77.56 & 85.05 & 98.68 & 27.78 & 54.83 \\
\checkmark & Binary & Gaussian   & 75.02 & 83.34 & 98.36 & 24.73 & 60.34 & 76.44 & 84.28 & 98.46 & 27.19 & 58.61 \\
\checkmark & Gaussian & Binary   & 76.90 & 84.85 & 98.59 & 26.46 & 56.46 & 78.17 & 85.63 & 98.71 & 28.82 & 54.81 \\
\rowcolor{lightgray!40}
\checkmark & Gaussian & Gaussian & \textbf{77.95} & \textbf{85.50} & \textbf{98.74} & \textbf{28.31} & \textbf{55.29} & \textbf{79.20} & \textbf{86.27} & \textbf{98.82} & \textbf{31.00} & \textbf{53.82} \\
\bottomrule[1.pt]
\end{tabular}
\end{table*}

The Prompt Guidance Generator (PGG) produces per-pixel guidance features that condition the Feature Enhancer. To understand the contribution of each guidance source, we conduct a systematic ablation starting from the configuration with no guidance inputs at all (Tab.~\ref{tab:pgg_prompt}, first row), which corresponds to the ``Without PGG'' variant in Tab.~\ref{tab:pgg_abs}. Adding the previous-stage prediction mask as the sole input improves IoU from 73.73 to 74.61 under LQ-3. However, at the first interaction no prior mask is available, leaving the enhancer without guidance and producing a lower-quality initial prediction that propagates through subsequent iterations (first-iteration IoU under LQ-3: 56.53 with mask-only vs.\ 62.98 with full prompts). Since point and box prompts encode complementary spatial cues, with points indicating error regions and boxes delineating object extent, we incorporate all three sources into PGG.

We further ablate how point and box prompts are spatially encoded: as hard binary indicator maps or as soft proximity maps generated with a Gaussian kernel. As reported in Tab.~\ref{tab:pgg_prompt}, incorporating prompt maps consistently improves over the mask-only baseline across all metrics (e.g., +3.34 IoU, +2.72 Dice, and $-$8.89 HD95 on average for the full model). The best configuration, Gaussian encoding for both point and box maps, achieves the highest results across all metrics, yielding 77.95 IoU and 55.29 HD95 under LQ-3.

\subsection{Comparison with LiteRobustSAM}

\begin{table}[tb]
\centering
\caption{\textbf{Comparison with LiteRobustSAM.} Both models use the pretrained LiteMedSAM and are trained under identical protocols. \textbf{Bold} denotes the best results.}
\setlength{\tabcolsep}{3.0pt}
\tiny
\renewcommand{\arraystretch}{1.2}
\begin{tabular}{ccccccccccccc}\toprule[1.pt]
     \multirow{2}{*}{{Method}} & \multicolumn{2}{c}{{LQ-3+}} & \multicolumn{2}{c}{{LQ-3}} &\multicolumn{2}{c}{{LQ-2}} & \multicolumn{2}{c}{{LQ-1}} & \multicolumn{2}{c}{{Clear}} &\multicolumn{2}{c}{{Average}} \\
\cmidrule(r){2-3}  \cmidrule(r){4-5} \cmidrule(r){6-7} \cmidrule(r){8-9}  \cmidrule(r){10-11} \cmidrule(r){12-13}
                               & IoU & Dice & IoU & Dice 
                            & IoU & Dice & IoU & Dice & IoU & Dice 
                            & IoU & Dice \\
\toprule[1.pt]
LiteRobustSAM        & 73.24 & 82.34 & 73.50 & 82.51 & 74.87 & 83.43 & 75.57 & 83.88 & 75.72 & 83.98 & 74.58 & 83.23 \\
\rowcolor{lightgray!40}
Lite\mName{}           & \textbf{77.22} & \textbf{85.05} & \textbf{77.95} & \textbf{85.50} & \textbf{79.71} & \textbf{86.57} & \textbf{80.51} & \textbf{87.10} & \textbf{80.61} & \textbf{87.14} & \textbf{79.20} & \textbf{86.27} \\
\bottomrule[1.pt]
\end{tabular}
\label{tab:robustmed}
\end{table}

A potential concern is whether the performance gains of Lite\mName{} over existing methods arise primarily from differences in training protocol rather than architectural design. To address this, we construct LiteRobustSAM, a lightweight variant that applies the RobustSAM architecture on top of the pretrained LiteMedSAM backbone, and train it under the exact same protocol as Lite\mName{} (same data, same augmentation, same number of epochs and iterative steps). As reported in Tab.~\ref{tab:robustmed}, Lite\mName{} consistently outperforms LiteRobustSAM across all degradation levels by approximately 4--5 points in IoU (e.g., 77.95 vs.\ 73.50 under LQ-3). This gap persists even on clean images (80.61 vs.\ 75.72 IoU), confirming that the gains stem from our architectural design, specifically the prompt-guided enhancement, multi-scale feature interaction, and foreground-focused supervision, rather than from training setup differences.

\subsection{Ablation on FR Loss Weight}

\begin{table*}[tb]
\centering
\caption{\textbf{Ablation on FR loss weight $\lambda_2$.} We vary the weight of the Foreground Reconstruction loss. \textbf{Bold} denotes the best results.}
\label{tab:loss_weight}
\setlength{\tabcolsep}{2.0pt}
\begin{tabular}{c|ccccc|ccccc}\toprule[1.pt]
\multirow{2}{*}{$\lambda_2$} & \multicolumn{5}{c|}{LQ-3} & \multicolumn{5}{c}{Average} \\
\cmidrule(r){2-6} \cmidrule(r){7-11}
 & IoU$\uparrow$ & Dice$\uparrow$ & PA$\uparrow$ & NSD$\uparrow$ & HD95$\downarrow$ & IoU$\uparrow$ & Dice$\uparrow$ & PA$\uparrow$ & NSD$\uparrow$ & HD95$\downarrow$ \\
\midrule
0  & 74.81 & 83.12 & 98.55 & 22.12 & 58.85 & 75.80 & 83.77 & 98.64 & 23.53 & 56.77 \\
1  & 76.40 & 83.97 & 98.71 & 25.79 & 58.92 & 77.73 & 84.88 & 98.81 & 28.41 & 55.56 \\
2  & 75.79 & 84.00 & 98.51 & 24.40 & 56.11 & 77.11 & 84.85 & 98.60 & 26.73 & 54.03 \\
\rowcolor{lightgray!40}
5  & \textbf{77.95} & \textbf{85.50} & \textbf{98.74} & \textbf{28.31} & \textbf{55.29} & \textbf{79.20} & \textbf{86.27} & \textbf{98.82} & \textbf{31.00} & \textbf{53.82} \\
10 & 75.53 & 83.67 & 98.42 & 24.56 & 56.37 & 76.57 & 84.47 & 98.50 & 26.60 & 55.80 \\
\bottomrule[1.pt]
\end{tabular}
\end{table*}

We ablate the weight $\lambda_2$ of the Foreground Reconstruction (FR) loss, which encourages the Feature Enhancer to reconstruct features within the foreground region. The Dice and Focal loss weights are fixed at 1 and 20, respectively, following prior work~\cite{sam,robustsam,glesam}. As reported in Tab.~\ref{tab:loss_weight}, setting $\lambda_2{=}0$ (i.e., no reconstruction supervision) yields the lowest performance (74.81 IoU, 22.12 NSD under LQ-3), confirming that the FR loss provides a meaningful learning signal for degradation-aware enhancement. Increasing $\lambda_2$ progressively improves results, with $\lambda_2{=}5$ achieving the best across all metrics (77.95 IoU on average). However, further increasing to $\lambda_2{=}10$ degrades performance, likely because an overly strong reconstruction objective shifts the optimization focus away from segmentation-relevant features. We therefore set $\lambda_2{=}5$ in all experiments.

\subsection{Ablation on Number of EBlocks per Stage}

\begin{table*}[tb]
\centering
\caption{\textbf{Ablation on the number of EBlocks per stage.} Performance on seen (ThinObject-5K, LVIS) and unseen (ECSSD, COCO) datasets under LQ-3 degradation with 5 interaction steps. \textbf{Bold} denotes the best results.}
\label{tab:eblock_ablation}
\scriptsize
\setlength{\tabcolsep}{1.5pt}
\renewcommand{\arraystretch}{1.2}
\begin{tabular}{c|c|ccc|ccc|ccc|ccc}\toprule[1.pt]
\multirow{2}{*}{$N$} & Added & \multicolumn{3}{c|}{ThinObject-5K} & \multicolumn{3}{c|}{LVIS} & \multicolumn{3}{c|}{ECSSD} & \multicolumn{3}{c}{COCO} \\
\cmidrule(r){3-5} \cmidrule(r){6-8} \cmidrule(r){9-11} \cmidrule(r){12-14}
 & Params (M) & IoU & Dice & NSD & IoU & Dice & NSD & IoU & Dice & NSD & IoU & Dice & NSD \\
\midrule
1  & 4.2 & 79.62 & 86.78 & 36.32 & 71.33 & 82.14 & 29.53 & 78.58 & 86.95 & 19.62 & 72.22 & 82.69 & 29.91 \\
3  & 8.0 & 80.05 & 87.22 & 37.00 & 72.46 & 83.04 & 30.37 & 78.97 & 87.29 & 20.04 & 72.74 & 83.00 & 30.23 \\
\rowcolor{lightgray!40}
5  & 11.7 & \textbf{81.27} & \textbf{88.09} & \textbf{38.38} & \textbf{73.30} & \textbf{83.46} & \textbf{31.80} & \textbf{79.91} & \textbf{87.89} & \textbf{20.90} & \textbf{73.56} & 83.64 & 31.25 \\
8  & 17.4 & 80.58 & 87.62 & 37.52 & 72.85 & 83.35 & 31.34 & 79.82 & 87.86 & 20.66 & 73.50 & \textbf{83.68} & \textbf{31.32}  \\
\bottomrule[1.pt]
\end{tabular}
\end{table*}

We ablate the number of Enhancement Blocks ($N$) per stage in the Feature Enhancer for the base \mName{} (ViT-B). As reported in Tab.~\ref{tab:eblock_ablation}, performance improves steadily from $N{=}1$ (4.2M added parameters) to $N{=}5$ (11.7M), with $N{=}5$ achieving the best results on the majority of metrics across all four datasets (e.g., 81.27 IoU on ThinObject-5K, 73.30 on LVIS). However, increasing to $N{=}8$ (17.4M) does not yield further gains; instead, performance slightly decreases on most datasets despite adding nearly 50\% more parameters. This suggests that five blocks per stage provide sufficient capacity for prompt-guided feature enhancement, and additional blocks lead to diminishing returns. We therefore adopt $N{=}5$ as the default configuration for \mName{}, striking a favorable balance between segmentation quality and parameter efficiency.

\section{Additional Experiments}
\label{supp:sec:expe}

\subsection{Performance on Clear Images}

\begin{table*}[tb]
\centering
\caption{\textbf{Performance on clear (undegraded) natural images} on seen (ThinObject-5K, LVIS) and unseen (ECSSD, COCO) datasets with 5 interaction steps. \textbf{Bold} and \underline{underline} indicate the best and second-best results, respectively.}
\label{tab:clear_natural}
\scriptsize
\setlength{\tabcolsep}{2.0pt}
\renewcommand{\arraystretch}{1.2}
\begin{tabular}{c|ccc|ccc|ccc|ccc}\toprule[1.pt]
\multirow{2}{*}{Method} & \multicolumn{3}{c|}{ThinObject-5K} & \multicolumn{3}{c|}{LVIS} & \multicolumn{3}{c|}{ECSSD} & \multicolumn{3}{c}{COCO} \\
\cmidrule(r){2-4} \cmidrule(r){5-7} \cmidrule(r){8-10} \cmidrule(r){11-13}
 & IoU & Dice & NSD & IoU & Dice & NSD & IoU & Dice & NSD & IoU & Dice & NSD \\
\midrule
SAM         & \underline{83.98} & \underline{89.93} & \underline{35.43} & \underline{79.95} & \underline{87.72} & \underline{41.53} & \underline{92.06} & \underline{95.49} & \underline{41.45} & \underline{79.83} & \underline{87.53} & \underline{41.20} \\
RobustSAM   & 58.75 & 69.72 & 27.22 & 37.39 & 47.11 & 15.16 & 65.56 & 75.27 & 29.01 & 39.91 & 49.99 & 16.00 \\
GleSAM      & 60.66 & 71.50 & 29.76 & 27.08 & 35.80 & 10.76 & 56.87 & 67.43 & 24.63 & 27.36 & 36.04 & 10.46 \\
\rowcolor{lightgray!40}
\mName      & \textbf{89.32} & \textbf{93.68} & \textbf{52.20} & \textbf{80.59} & \textbf{88.43} & \textbf{43.75} & \textbf{94.14} & \textbf{96.84} & \textbf{51.69} & \textbf{81.02} & \textbf{88.73} & \textbf{43.48} \\
\bottomrule[1.pt]
\end{tabular}
\end{table*}

Clear-image results for the medical domain (\dataName{}) were already reported in the main paper. Here, we complement that analysis by evaluating on the clear (undegraded) versions of all four LQ-Seg natural image datasets. As shown in Tab.~\ref{tab:clear_natural}, \mName{} not only maintains but improves upon SAM's performance on clean images, achieving gains of over 5 points in IoU on ThinObject-5K (89.32 vs.\ 83.98) and over 2 points on ECSSD (94.14 vs.\ 92.06). The NSD improvements are even more pronounced (e.g., 52.20 vs.\ 35.43 on ThinObject-5K), indicating substantially sharper boundary predictions. Together with the medical-domain results in the main paper, these findings confirm that our degradation-aware training does not sacrifice clean-image quality in either domain; instead, prompt-guided feature enhancement benefits segmentation regardless of whether the input is degraded.

\subsection{Additional Results on RobustSeg-style degradation}

\begin{table*}[tb]
\centering
\caption{\textbf{Zero-shot evaluation on RobustSeg-style degradation~\cite{robustsam}} on seen (ThinObject-5K, LVIS) and unseen (ECSSD, COCO) datasets. All methods are evaluated with 5 iterative prompt steps. \mName{} is not trained on this degradation type. \textbf{Bold} and \underline{underline} indicate the best and second-best results, respectively.}
\label{tab:robust_seg}
\scriptsize
\setlength{\tabcolsep}{2.0pt}
\renewcommand{\arraystretch}{1.2}
\begin{tabular}{c|ccc|ccc|ccc|ccc}\toprule[1.pt]
\multirow{2}{*}{Method} & \multicolumn{3}{c|}{ThinObject-5K} & \multicolumn{3}{c|}{LVIS} & \multicolumn{3}{c|}{ECSSD} & \multicolumn{3}{c}{COCO} \\
\cmidrule(r){2-4} \cmidrule(r){5-7} \cmidrule(r){8-10} \cmidrule(r){11-13}
 & IoU & Dice & NSD & IoU & Dice & NSD & IoU & Dice & NSD & IoU & Dice & NSD \\
\midrule
SAM         & 78.15 & 85.39 & 31.41 & 75.57 & 84.27 & 37.15 & \underline{87.33} & 92.37 & 33.55 & 76.46 & 85.04 & 36.80 \\
MoCE-SAM    & 77.52 & 85.24 & 25.81 & 74.36 & 83.54 & 33.62 & 85.57 & 91.12 & 30.96 & 74.29 & 83.40 & 33.20 \\
DCPT-SAM    & \underline{79.74} & \underline{86.80} & \underline{31.90} & \underline{76.51} & \underline{85.15} & \underline{37.28} & 87.30 & \underline{92.42} & \underline{33.60} & \underline{76.75} & \underline{85.33} & \underline{36.93} \\
RobustSAM   & 54.23 & 66.22 & 21.81 & 35.40 & 45.31 & 12.97 & 56.29 & 67.65 & 18.76 & 35.79 & 45.88 & 12.83 \\
GleSAM      & 55.99 & 67.28 & 23.27 & 24.19 & 32.73 & 8.48 & 52.52 & 63.91 & 17.96 & 24.68 & 33.31 & 8.77 \\
\rowcolor{lightgray!40}
\mName      & \textbf{86.98} & \textbf{92.15} & \textbf{47.74} & \textbf{78.15} & \textbf{86.79} & \textbf{40.09} & \textbf{90.41} & \textbf{94.62} & \textbf{41.18} & \textbf{78.91} & \textbf{87.34} & \textbf{39.74} \\
\bottomrule[1.pt]
\end{tabular}
\end{table*}

To evaluate generalization across degradation paradigms, we additionally benchmark \mName{} on the RobustSeg degradation pipeline introduced in RobustSAM~\cite{robustsam}. Unlike the LQ-Seg protocol~\cite{glesam} used in our main experiments, RobustSeg synthesizes 15 types of degradation spanning blur, noise, low-light conditions, and adverse weather. Crucially, \mName{} is \emph{not} trained on RobustSeg-style degradations, making this a zero-shot cross-degradation evaluation.
As reported in Tab.~\ref{tab:robust_seg}, \mName{} substantially outperforms all baselines on both seen (ThinObject-5K, LVIS) and unseen (ECSSD, COCO) datasets~\cite{thin,lvis,ecssd,coco}. Notably, RobustSAM, despite being trained on this degradation type, performs poorly under iterative evaluation. The cascade baselines (MoCE-SAM, DCPT-SAM) perform comparably to vanilla SAM, suggesting that generic restoration offers limited benefit when the degradation distribution shifts. These results confirm that our prompt-guided enhancement strategy generalizes robustly even to unseen degradation paradigms.

\subsection{Additional Results on Zero-shot Real-world Datasets}

To further validate the practical applicability of \mName{}, we evaluate on three real-world degraded image datasets that are entirely unseen during training. Unlike the synthetic degradation benchmarks above, these datasets capture naturally occurring image quality challenges, including low-light conditions, adverse weather, and underwater turbidity.

\subsubsection{Results on LIS Dataset}

\begin{table}[tb]
\centering
\caption{\textbf{Zero-shot evaluation on the LIS dataset~\cite{lis}.} All methods are evaluated with 5 iterative prompt steps. \textbf{Bold} indicates the best result.}
\label{tab:lis}
\setlength{\tabcolsep}{4pt}
\begin{tabular}{c|ccccc}\toprule[1.pt]
\multirow{2}{*}{Method} & \multicolumn{5}{c}{LIS} \\
\cmidrule(r){2-6}
& IoU $\uparrow$ & Dice $\uparrow$ & PA $\uparrow$ & NSD $\uparrow$ & HD95 $\downarrow$ \\
\midrule
SAM         & 78.14 & 85.98 & 97.57 & 33.37 & 49.97 \\
RobustSAM   & 41.85 & 52.07 & 73.77 & 11.99 & 325.98 \\
GleSAM      & 25.98 & 35.76 & 66.84 & 6.45 & 459.54 \\
\rowcolor{lightgray!40}
\mName      & \textbf{80.55} & \textbf{88.14} & \textbf{98.52} & \textbf{36.16} & \textbf{40.29} \\
\bottomrule[1.pt]
\end{tabular}
\end{table}

The LIS dataset~\cite{lis} contains 2,230 real-world low-light images with instance-level pixel-wise annotations covering diverse daily-life object categories in both indoor and outdoor scenes. As shown in Tab.~\ref{tab:lis}, \mName{} outperforms SAM by 2.4 points in IoU (80.55 vs.\ 78.14), 2.2 in Dice, and notably reduces HD95 from 49.97 to 40.29, indicating sharper boundary predictions under low-light conditions. \mName{} also improves NSD by nearly 3 points (36.16 vs.\ 33.37), further confirming better boundary quality.

\subsubsection{Results on WXSDO Dataset}

\begin{table}[tb]
\centering
\caption{\textbf{Zero-shot evaluation on the WXSDO dataset~\cite{wxsdo}.} All methods are evaluated with 5 iterative prompt steps. \textbf{Bold} indicates the best result.}
\label{tab:wxsdo}
\setlength{\tabcolsep}{4pt}
\begin{tabular}{c|ccccc}\toprule[1.pt]
\multirow{2}{*}{Method} & \multicolumn{5}{c}{WXSDO} \\
\cmidrule(r){2-6}
& IoU $\uparrow$ & Dice $\uparrow$ & PA $\uparrow$ & NSD $\uparrow$ & HD95 $\downarrow$ \\
\midrule
SAM         & 92.33 & 95.60 & 98.88 & 66.56 & 19.37 \\
RobustSAM   & 61.27 & 69.75 & 80.22 & 32.62 & 205.77 \\
GleSAM      & 39.01 & 48.24 & 69.26 & 17.68 & 371.91 \\
\rowcolor{lightgray!40}
\mName      & \textbf{93.18} & \textbf{96.31} & \textbf{99.48} & \textbf{68.90} & \textbf{17.22} \\
\bottomrule[1.pt]
\end{tabular}
\end{table}

The WXSDO dataset~\cite{wxsdo} targets salient object detection under adverse weather. We evaluate on its real-world subset, which contains 554 images captured under diverse weather conditions including rain, fog, snow, and varying lighting. As reported in Tab.~\ref{tab:wxsdo}, despite SAM already achieving strong performance (92.33 IoU), \mName{} still improves IoU to 93.18, Dice to 96.31, and reduces HD95 from 19.37 to 17.22. The NSD gain of over 2 points (68.90 vs.\ 66.56) further demonstrates that prompt-guided enhancement consistently benefits boundary-level accuracy even when overall segmentation quality is already high.

\subsubsection{Results on SMDD Dataset}

\begin{table}[tb]
\centering
\caption{\textbf{Zero-shot evaluation on the SMDD dataset~\cite{smdd}.} All methods are evaluated with 5 iterative prompt steps. \textbf{Bold} indicates the best result.}
\label{tab:smdd}
\setlength{\tabcolsep}{4pt}
\begin{tabular}{c|ccccc}\toprule[1.pt]
\multirow{2}{*}{Method} & \multicolumn{5}{c}{SMDD} \\
\cmidrule(r){2-6}
& IoU $\uparrow$ & Dice $\uparrow$ & PA $\uparrow$ & NSD $\uparrow$ & HD95 $\downarrow$ \\
\midrule
SAM         & 77.49 & 86.04 & 98.33 & 38.97 & 31.11 \\
RobustSAM   & 36.26 & 45.71 & 70.17 & 15.02 & 367.47 \\
GleSAM      & 22.09 & 29.54 & 62.55 & 10.16 & 492.79 \\
\rowcolor{lightgray!40}
\mName      & \textbf{79.36} & \textbf{87.66} & \textbf{99.32} & \textbf{40.51} & \textbf{20.82} \\
\bottomrule[1.pt]
\end{tabular}
\end{table}

The SMDD dataset~\cite{smdd} is the first publicly available underwater marine debris dataset, containing images captured in shallow-water environments with varying lighting and turbidity conditions. As shown in Tab.~\ref{tab:smdd}, \mName{} improves upon SAM by 1.9 points in IoU (79.36 vs.\ 77.49) and achieves the lowest HD95 of 20.82, a 33\% reduction compared to SAM (31.11). Across all three real-world datasets, RobustSAM and GleSAM consistently underperform SAM by large margins, which we attribute to their lack of iterative refinement support as discussed in the main paper.

\subsection{Backbone Comparison across Different Scales}

\begin{table*}[tb]
\centering
\caption{\textbf{Backbone comparison on seen natural image datasets.} Performance across three degradation levels on ThinObject-5K and LVIS with 5 interaction steps. \textbf{Bold} and \underline{underline} denote the best and second-best results within each backbone group.}
\label{tab:backbone_seen}
\scriptsize
\setlength{\tabcolsep}{1.5pt}
\renewcommand{\arraystretch}{1.2}
\begin{tabular}{c|cccccc|cccccc}\toprule[1.pt]
\multirow{3}{*}{Method} & \multicolumn{6}{c|}{ThinObject-5K} & \multicolumn{6}{c}{LVIS} \\
\cmidrule(r){2-7} \cmidrule(r){8-13}
 & \multicolumn{2}{c}{LQ-3} & \multicolumn{2}{c}{LQ-2} & \multicolumn{2}{c|}{LQ-1} & \multicolumn{2}{c}{LQ-3} & \multicolumn{2}{c}{LQ-2} & \multicolumn{2}{c}{LQ-1} \\
\cmidrule(r){2-3} \cmidrule(r){4-5} \cmidrule(r){6-7} \cmidrule(r){8-9} \cmidrule(r){10-11} \cmidrule(r){12-13}
 & IoU & Dice & IoU & Dice & IoU & Dice & IoU & Dice & IoU & Dice & IoU & Dice \\
\midrule
\multicolumn{13}{c}{\textit{Tiny-ViT Backbone}} \\
\midrule
MobileSAM       & \underline{66.69} & \underline{76.35} & \underline{70.45} & \underline{79.18} & \underline{74.63} & \underline{82.58} & \underline{63.46} & \underline{74.73} & \underline{65.53} & \underline{75.94} & \underline{70.29} & \underline{80.12} \\
\rowcolor{lightgray!40}
Mobile\mName    & \textbf{79.25} & \textbf{86.70} & \textbf{81.26} & \textbf{88.04} & \textbf{83.77} & \textbf{89.84} & \textbf{72.39} & \textbf{82.83} & \textbf{74.39} & \textbf{84.26} & \textbf{75.88} & \textbf{85.30} \\
\midrule
\multicolumn{13}{c}{\textit{ViT-L Backbone}} \\
\midrule
SAM-L           & \underline{72.16} & \underline{80.67} & \underline{77.19} & \underline{84.46} & \underline{80.18} & \underline{87.03} & \underline{65.36} & \underline{76.00} & \underline{69.77} & \underline{79.64} & \underline{72.80} & \underline{82.12} \\
RobustSAM-L     & 61.88 & 72.68 & 68.95 & 78.20 & 73.07 & 81.82 & 39.33 & 50.10 & 43.52 & 54.24 & 46.22 & 56.71 \\
GleSAM-L        & 54.63 & 65.40 & 55.13 & 65.30 & 58.86 & 68.70 & 27.87 & 37.80 & 28.56 & 37.81 & 28.03 & 37.26 \\
\rowcolor{lightgray!40}
\mName-L        & \textbf{81.41} & \textbf{88.14} & \textbf{85.45} & \textbf{91.08} & \textbf{86.06} & \textbf{91.50} & \textbf{72.99} & \textbf{83.13} & \textbf{75.88} & \textbf{85.17} & \textbf{77.46} & \textbf{86.34} \\
\bottomrule[1.pt]
\end{tabular}
\end{table*}

\begin{table*}[tb]
\centering
\caption{\textbf{Backbone comparison on unseen natural image datasets.} Performance across three degradation levels on ECSSD and COCO with 5 interaction steps. \textbf{Bold} and \underline{underline} denote the best and second-best results within each backbone group.}
\label{tab:backbone_unseen}
\scriptsize
\setlength{\tabcolsep}{1.5pt}
\renewcommand{\arraystretch}{1.2}
\begin{tabular}{c|cccccc|cccccc}\toprule[1.pt]
\multirow{3}{*}{Method} & \multicolumn{6}{c|}{ECSSD} & \multicolumn{6}{c}{COCO} \\
\cmidrule(r){2-7} \cmidrule(r){8-13}
 & \multicolumn{2}{c}{LQ-3} & \multicolumn{2}{c}{LQ-2} & \multicolumn{2}{c|}{LQ-1} & \multicolumn{2}{c}{LQ-3} & \multicolumn{2}{c}{LQ-2} & \multicolumn{2}{c}{LQ-1} \\
\cmidrule(r){2-3} \cmidrule(r){4-5} \cmidrule(r){6-7} \cmidrule(r){8-9} \cmidrule(r){10-11} \cmidrule(r){12-13}
 & IoU & Dice & IoU & Dice & IoU & Dice & IoU & Dice & IoU & Dice & IoU & Dice \\
\midrule
\multicolumn{13}{c}{\textit{Tiny-ViT Backbone}} \\
\midrule
MobileSAM       & \underline{70.67} & \underline{80.71} & \underline{76.27} & \underline{85.01} & \underline{80.77} & \underline{88.14} & \underline{64.85} & \underline{75.98} & \underline{68.32} & \underline{78.58} & \underline{70.89} & \underline{80.72} \\
\rowcolor{lightgray!40}
Mobile\mName    & \textbf{78.39} & \textbf{86.82} & \textbf{82.51} & \textbf{89.66} & \textbf{85.40} & \textbf{91.55} & \textbf{73.07} & \textbf{83.32} & \textbf{75.16} & \textbf{84.78} & \textbf{76.50} & \textbf{85.68} \\
\midrule
\multicolumn{13}{c}{\textit{ViT-L Backbone}} \\
\midrule
SAM-L           & \underline{72.70} & \underline{82.30} & \underline{78.93} & \underline{86.82} & \underline{83.32} & \underline{89.81} & \underline{65.66} & \underline{76.41} & \underline{70.27} & \underline{80.08} & \underline{72.67} & \underline{81.87} \\
RobustSAM-L     & 56.93 & 69.23 & 63.40 & 74.64 & 70.27 & 80.03 & 40.57 & 51.70 & 45.26 & 56.11 & 46.20 & 56.77 \\
GleSAM-L        & 46.56 & 59.74 & 46.54 & 58.93 & 54.63 & 65.68 & 26.96 & 36.67 & 28.98 & 38.46 & 28.96 & 38.26 \\
\rowcolor{lightgray!40}
\mName-L        & \textbf{80.19} & \textbf{88.11} & \textbf{84.72} & \textbf{91.16} & \textbf{87.23} & \textbf{92.74} & \textbf{73.38} & \textbf{83.43} & \textbf{76.27} & \textbf{85.48} & \textbf{77.85} & \textbf{86.61} \\
\bottomrule[1.pt]
\end{tabular}
\end{table*}

To verify that our framework generalizes across backbone capacities, we evaluate \mName{} with two additional ViT backbones beyond the ViT-B variant reported in the main paper: the lightweight Tiny-ViT~\cite{tiny_vit} from MobileSAM~\cite{mobile_sam} and the larger ViT-L from SAM~\cite{sam}. For the ViT-L backbone, we additionally compare against RobustSAM-L and GleSAM-L. Results across all LQ-Seg datasets and degradation levels are reported in Tabs.~\ref{tab:backbone_seen} and~\ref{tab:backbone_unseen}.

\mName{} consistently outperforms all baselines regardless of backbone scale. With the Tiny-ViT backbone, Mobile\mName{} surpasses MobileSAM by over 10 points in IoU under LQ-3 on both seen datasets (e.g., 79.25 vs.\ 66.69 on ThinObject-5K), demonstrating that even a lightweight backbone benefits substantially from prompt-guided enhancement. With ViT-L, \mName{}-L outperforms SAM-L by approximately 9 points in IoU under LQ-3, while RobustSAM-L and GleSAM-L again lag behind vanilla SAM-L. These results confirm that the performance gains of our framework are architecture-agnostic and scale consistently with backbone capacity.

\subsection{Extended Iterative Evaluation}

\begin{figure}[tb]
\centering
\includegraphics[width=1.0\linewidth]{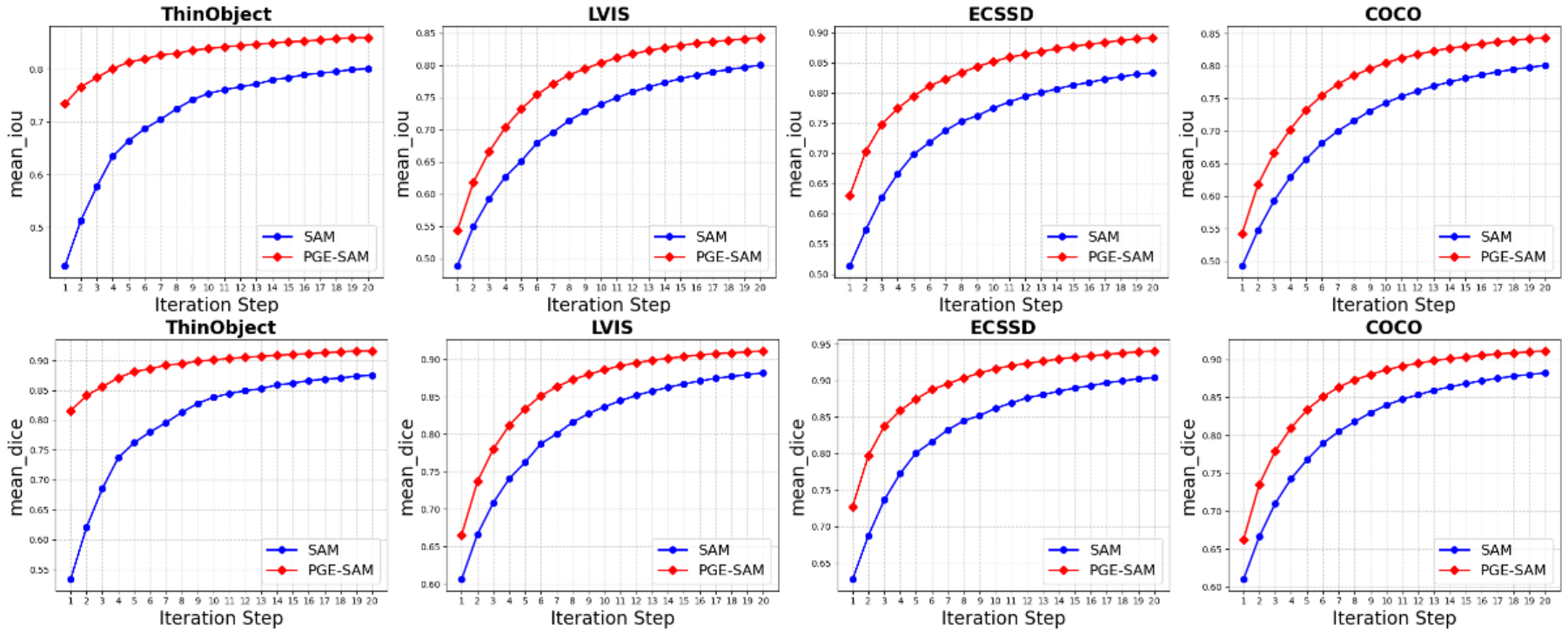}
   \caption{\textbf{Performance over 20 interaction steps on natural image datasets under LQ-3 degradation.} Mean IoU and Dice across ThinObject-5K, LVIS (seen) and ECSSD, COCO (unseen). \mName{} maintains consistent improvement throughout all 20 steps.}
\label{fig:stress_test_natural}
\end{figure}

\begin{figure}[tb]
\centering
\includegraphics[width=1.0\linewidth]{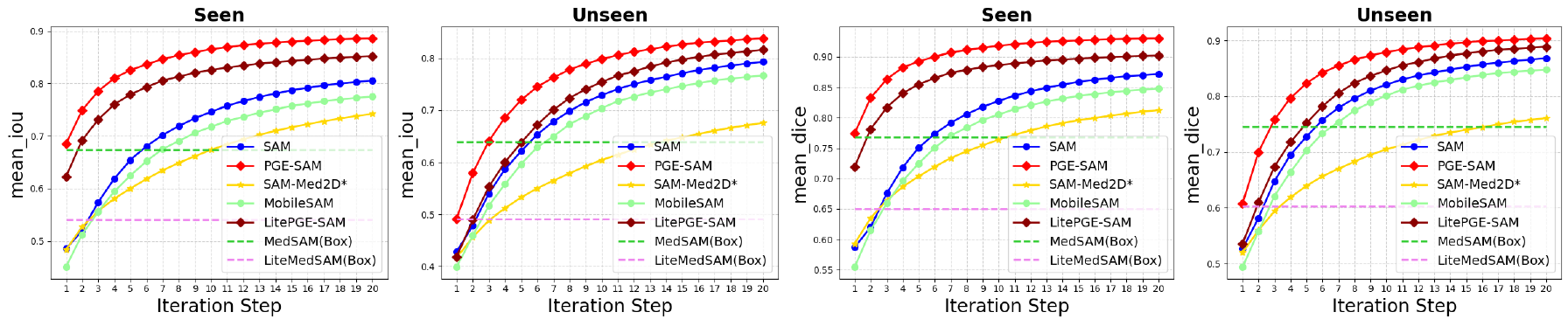}
   \caption{\textbf{Performance over 20 interaction steps on \dataName{} under LQ-3 degradation.} Mean IoU and Dice on seen and unseen splits. \mName{} and Lite\mName{} consistently improve with additional interactions, outperforming all baselines by a significant margin throughout the extended evaluation.}
\label{fig:stress_test_medical}
\end{figure}

Since \mName{} is trained with $K{=}5$ corrective iterations per sample, a natural question is whether the learned enhancement generalizes to longer interactive sessions where more than 5 refinement steps are required. To address this, we extend the evaluation to 20 interaction steps and report results under LQ-3 degradation on both domains. For natural images (Fig.~\ref{fig:stress_test_natural}), we compare \mName{} against SAM on all four LQ-Seg datasets. For medical images (Fig.~\ref{fig:stress_test_medical}), we include a broader set of baselines: SAM, MobileSAM, SAM-Med2D (without adapter)~\cite{sammed2d}, MedSAM~\cite{medsam}, LiteMedSAM, Lite\mName{}, and \mName{}, evaluated on both seen and unseen splits of \dataName{}. GleSAM and RobustSAM are excluded due to their consistently weak performance in iterative settings. As shown in both figures, \mName{} maintains steady improvement well beyond the 5-step training horizon and continues to outperform all baselines by a significant margin across all datasets. This confirms that our framework generalizes effectively to extended interactive sessions and is not bounded by the number of iterations used during training.

\subsection{Sensitivity to Initial Prompt Type}

\begin{table*}[tb]
\centering
\caption{\textbf{Performance with 3 random positive point initialization} across three degradation levels on seen and unseen datasets with 5 interaction steps. \textbf{Bold} and \underline{underline} denote best and second-best.}
\label{tab:3_point_prompt}
\scriptsize
\setlength{\tabcolsep}{2.0pt}
\renewcommand{\arraystretch}{1.2}
\begin{tabular}{c|cccccc|cccccc}\toprule[1.pt]
\multirow{3}{*}{Method} & \multicolumn{6}{c|}{ThinObject-5K (Seen)} & \multicolumn{6}{c}{LVIS (Seen)} \\
\cmidrule(r){2-7} \cmidrule(r){8-13}
 & \multicolumn{2}{c}{LQ-3} & \multicolumn{2}{c}{LQ-2} & \multicolumn{2}{c|}{LQ-1} & \multicolumn{2}{c}{LQ-3} & \multicolumn{2}{c}{LQ-2} & \multicolumn{2}{c}{LQ-1} \\
\cmidrule(r){2-3} \cmidrule(r){4-5} \cmidrule(r){6-7} \cmidrule(r){8-9} \cmidrule(r){10-11} \cmidrule(r){12-13}
 & IoU & Dice & IoU & Dice & IoU & Dice & IoU & Dice & IoU & Dice & IoU & Dice \\
\midrule
SAM         & \underline{72.12} & \underline{81.41} & \underline{76.18} & \underline{84.28} & \underline{79.27} & \underline{86.67} & \underline{62.18} & \underline{73.13} & \underline{65.88} & \underline{76.05} & \underline{69.03} & \underline{78.66} \\
RobustSAM   & 42.89 & 55.73 & 48.74 & 61.03 & 50.42 & 62.74 & 22.85 & 31.73 & 26.12 & 35.29 & 30.11 & 39.63 \\
GleSAM      & 61.29 & 72.57 & 67.27 & 77.39 & 61.49 & 72.73 & 28.15 & 38.04 & 32.77 & 43.04 & 28.81 & 38.89 \\
\rowcolor{lightgray!40}
\mName      & \textbf{81.09} & \textbf{88.03} & \textbf{84.11} & \textbf{90.31} & \textbf{85.23} & \textbf{90.99} & \textbf{73.70} & \textbf{84.00} & \textbf{75.90} & \textbf{85.45} & \textbf{76.87} & \textbf{86.10} \\
\toprule[1.pt]
 & \multicolumn{6}{c|}{ECSSD (Unseen)} & \multicolumn{6}{c}{COCO (Unseen)} \\
\midrule
SAM         & \underline{69.66} & \underline{80.11} & \underline{74.27} & \underline{83.38} & \underline{78.80} & \underline{86.65} & \underline{62.12} & \underline{73.07} & \underline{66.68} & \underline{76.72} & \underline{69.27} & \underline{78.80} \\
RobustSAM   & 36.31 & 49.81 & 39.98 & 53.31 & 44.91 & 57.76 & 23.82 & 32.72 & 28.32 & 37.74 & 31.18 & 40.90 \\
GleSAM      & 55.41 & 68.30 & 62.02 & 73.82 & 55.58 & 68.60 & 29.44 & 39.65 & 33.76 & 44.07 & 29.65 & 39.75 \\
\rowcolor{lightgray!40}
\mName      & \textbf{79.59} & \textbf{87.86} & \textbf{83.18} & \textbf{90.24} & \textbf{85.68} & \textbf{91.80} & \textbf{73.81} & \textbf{84.06} & \textbf{76.25} & \textbf{85.73} & \textbf{77.38} & \textbf{86.46} \\
\bottomrule[1.pt]
\end{tabular}
\end{table*}

\begin{table*}[tb]
\centering
\caption{\textbf{Performance with box-type initialization} across three degradation levels on seen and unseen datasets with 5 interaction steps. \textbf{Bold} and \underline{underline} denote best and second-best.}
\label{tab:box_prompt}
\scriptsize
\setlength{\tabcolsep}{2.0pt}
\renewcommand{\arraystretch}{1.2}
\begin{tabular}{c|cccccc|cccccc}\toprule[1.pt]
\multirow{3}{*}{Method} & \multicolumn{6}{c|}{ThinObject-5K (Seen)} & \multicolumn{6}{c}{LVIS (Seen)} \\
\cmidrule(r){2-7} \cmidrule(r){8-13}
 & \multicolumn{2}{c}{LQ-3} & \multicolumn{2}{c}{LQ-2} & \multicolumn{2}{c|}{LQ-1} & \multicolumn{2}{c}{LQ-3} & \multicolumn{2}{c}{LQ-2} & \multicolumn{2}{c}{LQ-1} \\
\cmidrule(r){2-3} \cmidrule(r){4-5} \cmidrule(r){6-7} \cmidrule(r){8-9} \cmidrule(r){10-11} \cmidrule(r){12-13}
 & IoU & Dice & IoU & Dice & IoU & Dice & IoU & Dice & IoU & Dice & IoU & Dice \\
\midrule
SAM         & \underline{62.79} & \underline{72.36} & \underline{69.60} & \underline{78.46} & \underline{72.66} & \underline{80.95} & \underline{69.79} & \underline{80.43} & \underline{72.86} & \underline{82.80} & \underline{74.82} & \underline{84.17} \\
RobustSAM   & 41.91 & 55.86 & 48.77 & 62.01 & 48.47 & 61.85 & 36.29 & 48.72 & 37.93 & 50.26 & 38.52 & 50.54 \\
GleSAM      & 39.74 & 52.82 & 42.31 & 55.47 & 42.80 & 55.81 & 13.49 & 20.38 & 15.31 & 22.69 & 15.77 & 23.19 \\
\rowcolor{lightgray!40}
\mName      & \textbf{81.72} & \textbf{88.34} & \textbf{84.82} & \textbf{90.76} & \textbf{85.18} & \textbf{90.85} & \textbf{75.02} & \textbf{84.61} & \textbf{77.30} & \textbf{86.29} & \textbf{78.23} & \textbf{86.87} \\
\toprule[1.pt]
 & \multicolumn{6}{c|}{ECSSD (Unseen)} & \multicolumn{6}{c}{COCO (Unseen)} \\
\midrule
SAM         & \underline{73.93} & \underline{83.36} & \underline{79.01} & \underline{87.08} & \underline{82.31} & \underline{89.32} & \underline{69.90} & \underline{80.60} & \underline{73.25} & \underline{83.08} & \underline{75.04} & \underline{84.32} \\
RobustSAM   & 47.38 & 62.44 & 50.42 & 64.86 & 53.99 & 67.97 & 35.70 & 48.12 & 38.30 & 50.58 & 40.24 & 52.50 \\
GleSAM      & 31.72 & 45.82 & 33.61 & 47.63 & 35.12 & 48.95 & 13.70 & 20.69 & 14.91 & 22.17 & 15.45 & 22.86 \\
\rowcolor{lightgray!40}
\mName      & \textbf{81.91} & \textbf{89.29} & \textbf{85.87} & \textbf{91.94} & \textbf{87.72} & \textbf{93.08} & \textbf{75.46} & \textbf{84.99} & \textbf{77.62} & \textbf{86.49} & \textbf{78.69} & \textbf{87.19} \\
\bottomrule[1.pt]
\end{tabular}
\end{table*}

All experiments in the main paper use a mixed prompt strategy that randomly selects between point and box prompts at the first interaction, following~\cite{scribbleprompt}. Here, we isolate the effect of the initial prompt type by evaluating under two fixed strategies: (1)~three random positive point prompts, as used in~\cite{glesam,robustsam}, and (2)~box-only prompts. As reported in Tabs.~\ref{tab:3_point_prompt} and~\ref{tab:box_prompt}, \mName{} consistently achieves the best performance across all datasets and degradation levels under both prompt types, demonstrating strong prompt-type robustness.

Notably, comparing the two tables reveals that GleSAM is highly sensitive to the initial prompt type. Under its preferred 3-point setting (Tab.~\ref{tab:3_point_prompt}), GleSAM achieves substantially higher results than under box prompts (Tab.~\ref{tab:box_prompt}) or the mixed strategy reported in the main paper (Tab.~\ref{tab:natural_image}). For instance, on LVIS under LQ-3, GleSAM's IoU drops from 28.15 with point prompts to just 13.49 with box prompts, a collapse of over 50\%. We attribute this to the fact that GleSAM was trained exclusively with positive point prompts and fine-tunes the mask decoder, leading to overfitting to this specific prompt configuration. In contrast, even under GleSAM's best-case setting, \mName{} outperforms it by a large margin across all datasets in an iterative manner. This analysis underscores the importance of prompt-type agnosticism in interactive segmentation, a property that \mName{} inherently possesses through its mixed-prompt training and iterative-aware design.

\subsection{Cross-Domain Generalization}

\begin{table}[tb]
\centering
\caption{\textbf{Cross-domain evaluation on \dataName{}.} $^\star$ denotes the \mName{} variant trained only on natural image data. \textbf{Bold} denotes the best results.}
\setlength{\tabcolsep}{3.0pt}
\tiny
\renewcommand{\arraystretch}{1.2}
\begin{tabular}{ccccccccccccc}\toprule[1.pt]
     \multirow{2}{*}{{Method}} & \multicolumn{2}{c}{{LQ-3+}} & \multicolumn{2}{c}{{LQ-3}} &\multicolumn{2}{c}{{LQ-2}} & \multicolumn{2}{c}{{LQ-1}} & \multicolumn{2}{c}{{Clear}} &\multicolumn{2}{c}{{Average}} \\
\cmidrule(r){2-3}  \cmidrule(r){4-5} \cmidrule(r){6-7} \cmidrule(r){8-9}  \cmidrule(r){10-11} \cmidrule(r){12-13}
                               & IoU & Dice & IoU & Dice 
                            & IoU & Dice & IoU & Dice & IoU & Dice 
                            & IoU & Dice \\
\midrule
SAM & 63.14 & 73.33 & 65.24 & 74.90 & 68.21 & 77.20 & 69.77 & 78.42 & 70.00 & 78.45 & 67.27 & 76.46 \\
\rowcolor{lightgray!40}
\mName{}$^\star$          & \textbf{70.03} & \textbf{80.14} & \textbf{70.08} & \textbf{80.03} & \textbf{72.28} & \textbf{81.54} & \textbf{72.81} & \textbf{81.88} & \textbf{73.11} & \textbf{82.10} & \textbf{71.66} & \textbf{81.14} \\
\bottomrule[1.pt]
\end{tabular}
\label{tab:cross_domain}
\end{table}

To evaluate cross-domain generalization, we take the \mName{} variant trained exclusively on natural images (ThinObject-5K, MSRA-10K, LVIS) using the SAM (ViT-B) backbone and directly evaluate it on the medical \dataName{} dataset without any medical-domain fine-tuning. This represents a fully zero-shot cross-domain setting, as neither SAM nor this variant of \mName{} has been trained on medical data. As shown in Tab.~\ref{tab:cross_domain}, \mName{}$^\star$ improves over SAM across all degradation levels (e.g., 70.08 vs.\ 65.24 under LQ-3) and notably by 3.1 points even on clean images (73.11 vs.\ 70.00). This demonstrates that the feature enhancement learned by \mName{} is not domain-specific: the prompt-guided enhancement module acquires general-purpose representations that transfer across visual domains, improving both degradation robustness and baseline segmentation quality.

\subsection{Comparison with VLM-Based Methods}

\begin{figure}[ht]
\centering
\includegraphics[width=1.0\textwidth]{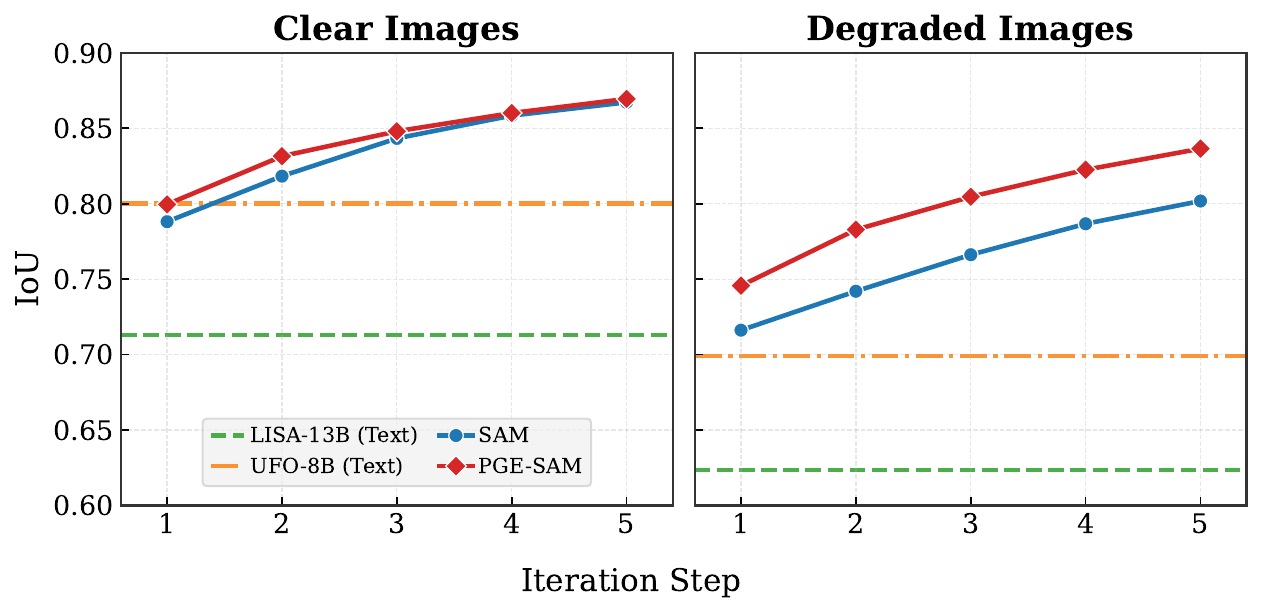}
\caption{Comparison with VLM-based methods on RefCOCO.}
\label{fig:vlm_comparison}
\end{figure}

We conduct an additional experiment to compare \mName{} with recent VLM-based methods \cite{ufo,lisa} on the RefCOCO dataset \cite{refcoco}, with results presented in Fig.~\ref{fig:vlm_comparison}. As shown, despite achieving competitive performance on clean images, the performance of both UFO and LISA drops significantly on their degraded counterparts. Furthermore, these VLM-based approaches lack an iterative mechanism, preventing them from progressively refining their predictions across interaction steps.

\subsection{Bounding-Box Robustness}

\begin{table}[ht]
\centering
\setlength{\tabcolsep}{2pt}
\caption{Bounding-box prompt robustness under the BREPS protocol (IoU, first iteration). T/M/X denote tight/min/max box prompts.}
\begin{tabular}{l|ccc|ccc}\toprule[1.pt]
 & \multicolumn{3}{c|}{ECSSD-Deg} & \multicolumn{3}{c}{ECSSD-Clr} \\
 & T & M & X & T & M & X \\
\midrule
SAM & 70.01 & 47.01 & 74.75 & 87.32 & 41.96 & 90.55 \\
\rowcolor{lightgray!40}
PGE-SAM & \textbf{75.24} & \textbf{54.84} & \textbf{78.99} & \textbf{91.53} & \textbf{59.31} & \textbf{93.85} \\
\bottomrule[1.pt]
\end{tabular}
\label{tab:breps}
\end{table}

In addition to robustness against image degradation, we evaluate the model's resilience to variations in user input using the BREPS \cite{breps} evaluation protocol. Specifically, we report performance using tight, min, and max bounding box prompts to simulate natural variations in user annotations. As reported in Tab.~\ref{tab:breps}, \mName{} is substantially more robust to user input variability than the baseline SAM in both clear and degraded scenarios. We attribute this improvement to our training pipeline, which directly simulates noisy box prompts during training. These results further demonstrate the robustness of \mName{} to challenges in both image quality and user interaction.

\section{Implementation Details}
\label{supp:sec:imple_detail}

We maintain two separate training configurations depending on the target domain. \textbf{For the medical image domain}, we use the pretrained MedSAM (ViT-B) and LiteMedSAM (Tiny-ViT) backbones for \mName{} and Lite\mName{}, respectively. Models are trained on the training split of our proposed \dataName{} dataset and evaluated on its seen and unseen validation splits. \textbf{For the natural image domain}, we use the pretrained SAM backbones (ViT-B for \mName{}, ViT-L for \mName{}-L). Training is conducted on three datasets: LVIS~\cite{lvis}, MSRA-10K~\cite{msra}, and ThinObject-5K~\cite{thin}, comprising approximately 30K image-mask pairs in total, with multi-level degradation synthesized as random combinations of common degradation models. The seen evaluation sets consist of ThinObject-5K and LVIS evaluation splits, while ECSSD~\cite{ecssd} and COCO-val~\cite{coco} serve as unseen sets. All models are evaluated across three degradation levels following the protocol of GleSAM~\cite{glesam}.

Tab.~\ref{tab:hyperparameters} summarizes the full training configurations for all three model variants. Lite\mName{} is built on the LiteMedSAM backbone (Tiny-ViT, $256{\times}256$ input), \mName{} on MedSAM or SAM (ViT-B, $1024{\times}1024$ input), and \mName{}-L on SAM (ViT-L, $1024{\times}1024$ input). In all cases, the Image Encoder is kept entirely frozen to preserve the pretrained representations; only the Feature Enhancer, Mask Decoder, and (for Lite\mName{}, because LiteMedSAM fine-tunes the Prompt Encoder for box-specific prompts) the Prompt Encoder are updated during training. The Feature Enhancer uses three sequential stages in all variants, with 1 EBlock per stage for Lite\mName{}, 5 for \mName{}, and 8 for \mName{}-L, yielding 4.05M, 11.75M, and 19.47M added parameters, respectively. All models are trained for 36 epochs on 4 NVIDIA A100 GPUs using AdamW with a base learning rate of $5{\times}10^{-4}$ and weight decay of $1{\times}10^{-2}$. Due to the larger input resolution and model capacity, per-GPU batch size is set to 16 for Lite\mName{} (total 64), 2 for \mName{} (total 8), and 1 for \mName{}-L (total 4). Automatic Mixed Precision (AMP) is enabled for all variants to reduce memory consumption. At each training iteration, we simulate $K{=}5$ interactive steps to align with the iterative evaluation protocol. The loss weights are shared across all variants: Dice loss with unit weight, Focal loss with $\lambda_1{=}20$, and FR loss with $\lambda_2{=}5$.

\begin{table}[tb]
\caption{\textbf{Training configurations for Lite\mName{}, \mName{}, and \mName{}-L.}}
\begin{center}
\scriptsize
\setlength{\tabcolsep}{4pt}
\begin{tabular}{l|ccc}
\toprule
\textbf{Setting} & \textbf{Lite\mName{}} & \textbf{\mName{}} & \textbf{\mName{}-L} \\
\midrule
\multicolumn{4}{c}{\textit{Base Model}} \\
\midrule
Pretrained Backbone & LiteMedSAM (Tiny-ViT) & MedSAM / SAM (ViT-B) & SAM (ViT-L) \\
Backbone Status & Frozen & Frozen & Frozen \\
Prompt Encoder Status & Fine-tuned & Frozen & Frozen \\
Input Resolution & $256 \times 256$ & $1024 \times 1024$ & $1024 \times 1024$ \\
Total Parameters (M) & 13.84 & 105.49 & 329.95 \\
\midrule
\multicolumn{4}{c}{\textit{Feature Enhancer Architecture}} \\
\midrule
Enhancement Stages & 3 & 3 & 3 \\
EBlocks per Stage ($N$) & 1 & 5 & 8 \\
Feature Dimension ($C$) & 256 & 256 & 256 \\
\midrule
\multicolumn{4}{c}{\textit{Training}} \\
\midrule
Optimizer & AdamW & AdamW & AdamW \\
Base Learning Rate & $5 \times 10^{-4}$ & $5 \times 10^{-4}$ & $5 \times 10^{-4}$ \\
Weight Decay & $1 \times 10^{-2}$ & $1 \times 10^{-2}$ & $1 \times 10^{-2}$ \\
Epochs & 36 & 36 & 36 \\
Batch Size (per GPU) & 16 & 2 & 1 \\
GPUs & 4 $\times$ A100 & 4 $\times$ A100 & 4 $\times$ A100 \\
Total Batch Size & 64 & 8 & 4 \\
Iterative Steps ($K$) & 5 & 5 & 5 \\
Using AMP & Yes & Yes & Yes \\
\midrule
\multicolumn{4}{c}{\textit{Loss Weights}} \\
\midrule
Dice Loss & 1 & 1 & 1 \\
$\lambda_1$ (Focal Loss) & 20 & 20 & 20 \\
$\lambda_2$ (FR Loss) & 5 & 5 & 5 \\
\bottomrule
\end{tabular}
\end{center}
\label{tab:hyperparameters}
\end{table}

\section{Dataset Processing and Degradation Simulation}
\label{supp:sec:dataset_processing}

This section provides comprehensive details on the dataset collection, degradation simulation framework, and implementation specifications summarized in Section \ref{sec:deg_med_dataset} of the main paper.

\subsection{Dataset Summary}
\label{supp:dataset_details}

Table~\ref{supp:tab:dataset_summary} summarizes all 17 medical image datasets used in this study, including imaging modality, anatomical region, sample counts, and train/test split information.
\begin{table}[htbp]
\centering
\caption{Summary of medical image datasets using augmented sample counts. Seen datasets use the processed train/validation splits available on disk. Unseen datasets are held out entirely for generalization evaluation.}
\label{supp:tab:dataset_summary}
\resizebox{\textwidth}{!}{%
\begin{tabular}{llllr}
\toprule
\textbf{Dataset} & \textbf{Modality} & \textbf{Anatomy/Task} & \textbf{Ref} & \textbf{\# of images} \\
\midrule
\multicolumn{5}{l}{\textit{Seen Datasets}} \\
\midrule
abdominalUS & Ultrasound & Abdominal organs & \cite{vitale2020improving} & 597 \\
BUSI & Ultrasound & Breast lesions & \cite{al2020dataset} & 355 \\
DDTI & Ultrasound & Thyroid nodules & \cite{pedraza2015open} & 422 \\
FLARE & CT & Abdominal organs & \cite{ma2022fast} & 4,543 \\
ISIC & Dermoscopy & Skin lesions & \cite{codella2018skin} & 3,694 \\
MMOTU & Ultrasound & Ovarian tumors & \cite{zhao2022mmotu} & 1,469 \\
NerveUS & Ultrasound & Nerve structures & \cite{ultrasound-nerve-segmentation} & 2,323 \\
MuscleUS & Ultrasound & Muscle tissue & \cite{marzola2021deep} & 4,564 \\
BraTS & MRI & Brain tumors & \cite{baid2021rsna} & 6,498 \\
ToothSeg & Dental CT/X-ray & Tooth segmentation & \cite{silva2018automatic} & 594 \\
PAXRay & X-ray & Chest anatomy & \cite{seibold2022detailed} & 14,753 \\
BBBC038 & Microscopy & Cell nuclei & \cite{caicedo2019nucleus} & 841 \\
\midrule
\multicolumn{4}{l}{\textbf{Total (Seen)}} & \textbf{40,653} \\
\midrule
\multicolumn{5}{l}{\textit{Unseen Datasets (Held-Out for Generalization Evaluation)}} \\
\midrule
CVC-ClinicDB & Endoscopy & Polyp segmentation & \cite{bernal2015wm} & 612 \\
Kvasir-SEG & Endoscopy & Polyp segmentation & \cite{jha2019kvasir} & 1,000 \\
LGG & MRI & Low-grade glioma & \cite{buda2019association} & 1,373 \\
LungRads & CT & Lung screening/segmentation & \cite{nam2024lung} & 972 \\
CHAOS & CT/MRI & Abdominal organs & \cite{kavur2021chaos} & 197 \\
\midrule
\multicolumn{4}{l}{\textbf{Total (Unseen)}} & \textbf{4,154} \\
\multicolumn{4}{l}{\textbf{Total (All)}} & \textbf{44,807} \\
\bottomrule
\end{tabular}%
}
\end{table}

\vspace{2pt}
\subsection{Degradation Specifications}
\label{supp:degradation_specs}

\subsubsection{General Degradations}

Table~\ref{supp:tab:general_degradations} describes the eight general degradation types applied across all medical imaging modalities, along with their clinical relevance.

\begin{table}[htbp]
\centering
\caption{General degradation types applied across all medical imaging modalities.}
\label{supp:tab:general_degradations}
\begin{tabular}{lp{9cm}}
\toprule
\textbf{Degradation} & \textbf{Description \& Clinical Relevance} \\
\midrule
Gaussian Noise & Additive noise simulating electronic sensor noise and thermal fluctuations in imaging equipment. \\
Salt \& Pepper & Impulse noise simulating transmission errors, defective sensor pixels, or analog-to-digital conversion artifacts. \\
Motion Blur & Directional blur simulating patient movement during image acquisition, common in modalities with longer exposure times. \\
Gaussian Blur & Isotropic blur simulating defocus, poor resolution, or suboptimal imaging parameters. \\
Brightness/Contrast & Intensity variations simulating inconsistent illumination, exposure settings, or display calibration issues. \\
Gamma Correction & Non-linear intensity mapping simulating display gamma variations and image post-processing artifacts. \\
JPEG Compression & Block artifacts from lossy compression, common in archived or transmitted medical images. \\
Glare & Localized bright spots simulating specular reflections or light source artifacts. \\
\bottomrule
\end{tabular}
\end{table}

\subsubsection{Domain-Specific Degradations}

Medical images exhibit modality-specific artifacts arising from the underlying physics of each imaging technique. Table~\ref{supp:tab:domain_degradations} summarizes the domain-specific degradations implemented for each modality.

\begin{table}[tb]
\centering
\caption{Domain-specific degradations categorized by imaging modality.}
\label{supp:tab:domain_degradations}
\begin{tabular}{llp{7cm}}
\toprule
\textbf{Modality} & \textbf{Degradation} & \textbf{Description} \\
\midrule
\multirow{6}{*}{Ultrasound} 
& Speckle Noise & Multiplicative granular noise from coherent wave interference patterns characteristic of ultrasound imaging.~\cite{hiremath2013speckle, duarte2020speckle} \\
& Shadow & Acoustic shadowing caused by highly reflective structures such as bones, commonly observed in ultrasound imaging.~\cite{app11031127} \\
& Bias Field & Smooth intensity inhomogeneity from non-uniform ultrasound beam propagation.~\cite{jindal2022deep} \\
& Vignette & Peripheral intensity falloff due to transducer geometry and beam characteristics. \\
& Reflection & Specular highlights from probe contact gel or strongly reflective anatomical boundaries. \\
& Dead Pixels & Local sensor dropouts or missing scan-line artifacts that suppress small image regions. \\
\midrule
\multirow{3}{*}{MRI} 
& Bias Field & Low-frequency intensity variation from RF coil inhomogeneity and B1 field non-uniformity.~\cite{jindal2022deep} \\
& Ringing & Gibbs artifacts (oscillations near high-contrast edges) from k-space truncation during acquisition.~\cite{block2008suppression} \\
& Ghosting & Shifted duplicate copies of anatomy from periodic motion or blood flow during acquisition.~\cite{reeder1997quantification} \\
\midrule
\multirow{2}{*}{CT} 
& Ring Artifacts & Concentric ring patterns from detector element miscalibration or defects.~\cite{barrett2004artifacts,triche2019recognizing} \\
& Beam Hardening & Cupping artifacts and dark streaks from preferential absorption of low-energy photons in polychromatic X-ray beams.~\cite{barrett2004artifacts, triche2019recognizing} \\

\midrule
X-ray & Grid Banding & Periodic intensity variations from anti-scatter grid misalignment or Moir\'{e} patterns.~\cite{triche2019recognizing, curry1990christensen} \\
\midrule
Dental CT/X-ray
& Shadow & Local attenuation artifacts caused by dense dental structures and acquisition geometry. \\
& Vignette & Peripheral intensity falloff caused by field-of-view and reconstruction effects. \\
\midrule
\multirow{3}{*}{Endoscopy} 
& Reflection & Specular highlights from wet mucosal surfaces and endoscope illumination.~\cite{saken2021impact} \\
& Vignette & Peripheral darkening due to endoscope illumination falloff. \\
& Shadow & Local under-exposure caused by folds, occlusion, or uneven illumination. \\
\midrule
\multirow{3}{*}{Dermoscopy}
& Ghosting & Duplicate or shifted structures caused by acquisition or motion artifacts. \\
& Dead Pixels & Local sensor defects that create missing or corrupted pixels. \\
& Reflection & Specular highlights from skin surface illumination. \\
\midrule
\multirow{2}{*}{Microscopy}
& Fog & Low-contrast haze caused by defocus, scattering, or slide preparation artifacts. \\
& Reflection & Bright local artifacts from slide illumination and optical interfaces. \\
\bottomrule
\end{tabular}
\end{table}

\subsubsection{Dataset-to-Degradation Mapping}

Table~\ref{supp:tab:dataset_degradation_mapping} presents the mapping between each dataset and its corresponding domain-specific degradations applied at the LQ3+ level.

\begin{table}[htbp]
\centering
\caption{Dataset-to-degradation mapping for domain-specific artifacts (LQ3+ level).}
\label{supp:tab:dataset_degradation_mapping}
\begin{tabular}{llll}
\toprule
\textbf{Dataset} & \textbf{Modality} & \textbf{Domain-Specific Degradations} & \textbf{Status} \\
\midrule
abdominalUS~\cite{vitale2020improving} & Ultrasound & Speckle Noise, Shadow, Vignette & Seen \\
BUSI~\cite{al2020dataset} & Ultrasound & Speckle Noise, Shadow, Reflection & Seen \\
DDTI~\cite{pedraza2015open} & Ultrasound & Speckle Noise, Bias Field, Shadow & Seen \\
NerveUS~\cite{ultrasound-nerve-segmentation} & Ultrasound & Speckle Noise, Vignette, Shadow & Seen \\
MuscleUS~\cite{marzola2021deep} & Ultrasound & Speckle Noise, Shadow, Bias Field & Seen \\
MMOTU~\cite{zhao2022mmotu} & Ultrasound & Dead Pixels & Seen \\
BraTS~\cite{baid2021rsna} & MRI & Bias Field, Ringing, Ghosting & Seen \\
FLARE~\cite{ma2022fast} & CT & Ring Artifacts, Beam Hardening & Seen \\
ToothSeg~\cite{silva2018automatic} & Dental CT & Beam Hardening, Shadow, Vignette & Seen \\
ISIC~\cite{codella2018skin} & Dermoscopy & Ghosting, Dead Pixels, Reflection & Seen \\
PAXRay~\cite{seibold2022detailed} & X-ray & Grid Banding & Seen \\
BBBC038~\cite{caicedo2019nucleus} & Microscopy & Fog, Reflection & Seen \\
\midrule
CHAOS~\cite{kavur2021chaos} & CT/MRI & Ring Artifacts, Bias Field & Unseen \\
Kvasir-SEG~\cite{jha2019kvasir} & Endoscopy & Reflection, Vignette, Shadow & Unseen \\
CVC-ClinicDB~\cite{bernal2015wm} & Endoscopy & Reflection, Vignette, Shadow & Unseen \\
LGG~\cite{buda2019association} & MRI & Bias Field, Ringing, Ghosting & Unseen \\
LungRads~\cite{nam2024lung} & CT & Ring Artifacts, Beam Hardening & Unseen \\
\bottomrule
\end{tabular}
\end{table}

\begin{figure}[t]
\centering

\includegraphics[width=1.0\textwidth]{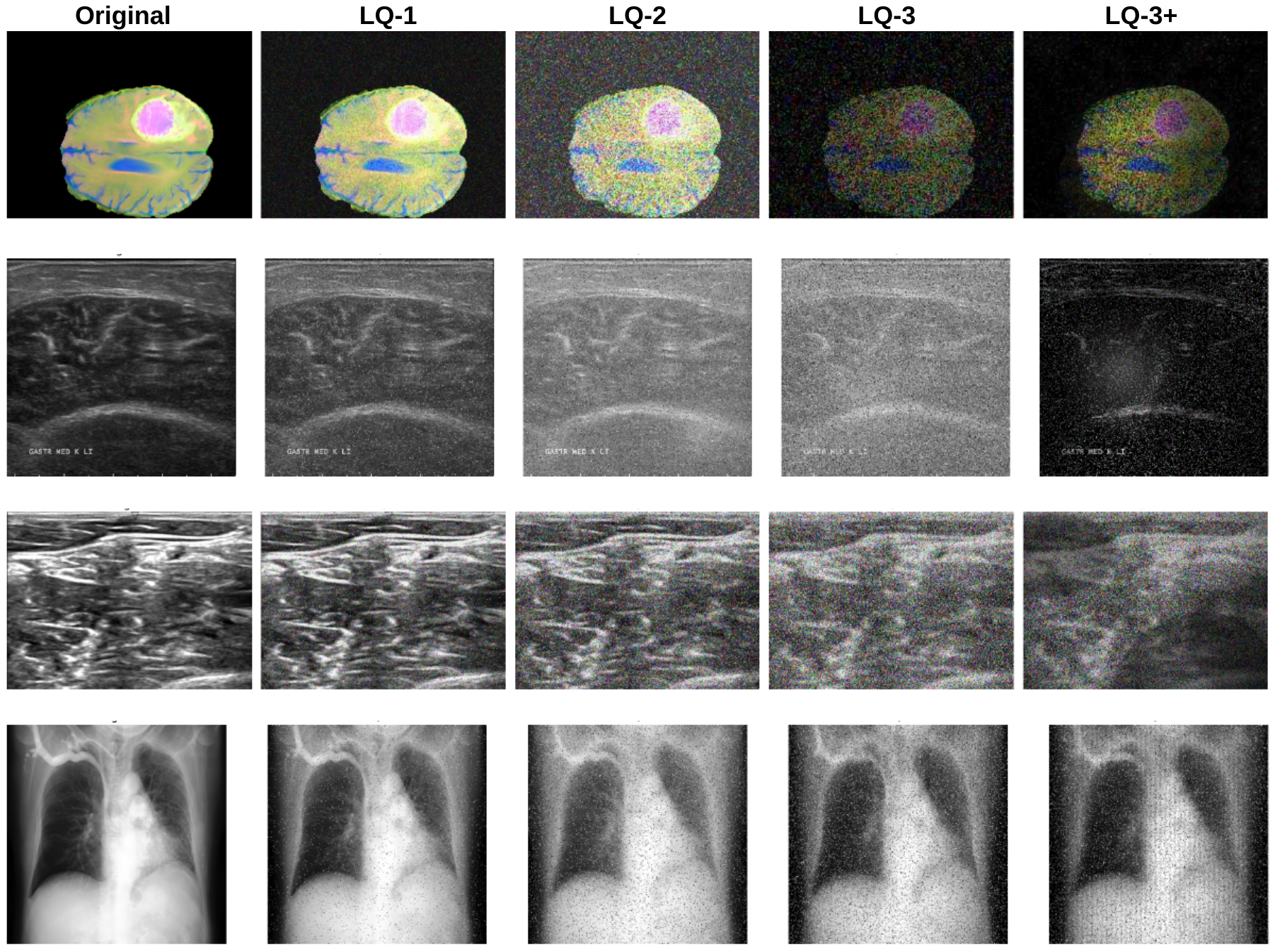}
\caption{Examples of degraded images sampled from different datasets.}
\label{fig:sampled_degraded_images}
\end{figure}

\subsection{Degradation Parameter Ranges}
\label{supp:param_specs}

The intensity of each degradation is controlled by level-specific parameters. Table~\ref{supp:tab:degradation_params} details the parameter ranges for general degradations at each quality level. Parameters are uniformly sampled from the specified ranges during augmentation.

\begin{table}[htbp]
\centering
\caption{Parameter ranges for general degradations at each quality level.}
\label{supp:tab:degradation_params}
\resizebox{\textwidth}{!}{%
\begin{tabular}{llccc}
\toprule
\textbf{Degradation} & \textbf{Parameter} & \textbf{LQ1 (Light)} & \textbf{LQ2 (Medium)} & \textbf{LQ3 (Heavy)} \\
\midrule
Gaussian Noise & noise\_factor & $[0.05, 0.15]$ & $[0.15, 0.30]$ & $[0.30, 0.50]$ \\
\midrule
\multirow{2}{*}{Salt \& Pepper} & salt\_prob & $[0.01, 0.03]$ & $[0.03, 0.06]$ & $[0.06, 0.12]$ \\
& pepper\_prob & $[0.01, 0.03]$ & $[0.03, 0.06]$ & $[0.06, 0.12]$ \\
\midrule
Motion Blur & kernel\_size (px) & $[3, 7]$ & $[7, 15]$ & $[15, 25]$ \\
\midrule
Gaussian Blur & $\sigma$ & $[0.5, 1.5]$ & $[1.5, 2.5]$ & $[2.5, 4.0]$ \\
\midrule
\multirow{2}{*}{Brightness/Contrast} & brightness\_factor & $[\pm 0.05, \pm 0.15]$ & $[\pm 0.15, \pm 0.30]$ & $[\pm 0.30, \pm 0.50]$ \\
& contrast\_factor & $[\pm 0.05, \pm 0.15]$ & $[\pm 0.15, \pm 0.30]$ & $[\pm 0.30, \pm 0.50]$ \\
\midrule
Gamma Correction & $\gamma$ & $[0.8, 1.2]$ & $[0.6, 1.6]$ & $[0.4, 2.5]$ \\
\midrule
JPEG Compression & quality & $[70, 90]$ & $[45, 70]$ & $[15, 45]$ \\
\midrule
\multirow{2}{*}{Glare} & intensity & $[0.05, 0.15]$ & $[0.15, 0.30]$ & $[0.30, 0.50]$ \\
& size (ratio) & $[0.1, 0.2]$ & $[0.2, 0.35]$ & $[0.35, 0.50]$ \\
\bottomrule
\end{tabular}%
}
\end{table}

\subsection{Evaluation Protocol}
\label{supp:subsec:evaluation_protocol}

We adopt a two-tier evaluation protocol to assess both in-domain performance and cross-domain generalization. In total, the \dataName{} benchmark comprises 44,807 images with multiple masks: 36,033 from 12 seen datasets used for training and 4,620 from 6 of these datasets reserved for seen validation, and 4,154 from 5 entirely held-out unseen datasets.

\paragraph{\textbf{Seen domain evaluation.}} From the 12 seen training datasets, we select the validation splits of 6 representative datasets to form the seen evaluation set, covering a diverse range of modalities. This split measures the model's ability to handle degradations similar to those encountered during training, representing interpolation performance within the learned distribution.

\paragraph{\textbf{Unseen domain evaluation.}} Models are additionally evaluated on 5 entirely held-out datasets (4,154 images) spanning CT, MRI, and endoscopy modalities. These datasets were never seen during training in any form, and they cover anatomical regions and acquisition protocols not fully represented in the seen datasets. This measures extrapolation performance beyond the training distribution.

All evaluations are conducted across five image quality levels: Clear, LQ-1, LQ-2, LQ-3, and LQ-3+, to comprehensively assess robustness under progressively severe degradation. Fig.~\ref{fig:sampled_degraded_images} illustrates sample images with different degradation levels from our proposed \dataName{} dataset.

\subsection{Final Dataset Statistics}
\label{supp:subsec:dataset_stats}

Table~\ref{supp:tab:dataset_summary} summarizes the final dataset statistics after augmentation. For each original image, five versions are generated: one clean image and four degraded variants (LQ1, LQ2, LQ3, LQ3+), resulting in a 5$\times$ expansion of the effective training data.
\subsection{Dataset Descriptions}
\label{supp:subsec:dataset_descriptions}

This section provides brief descriptions of each dataset used in our study.

\subsubsection{Seen Datasets}

\begin{itemize}
    \item \textbf{abdominalUS}~\cite{vitale2020improving}: Ultrasound images of abdominal organs including liver, kidney, and spleen with corresponding segmentation masks.
    
    \item \textbf{BUSI}~\cite{al2020dataset}: Breast Ultrasound Images dataset containing normal, benign, and malignant breast ultrasound images with ground truth lesion segmentations.
    
    \item \textbf{DDTI}~\cite{pedraza2015open}: Digital Database of Thyroid Imaging containing ultrasound images of thyroid nodules with expert-annotated segmentation masks.
    
    \item \textbf{NerveUS}~\cite{ultrasound-nerve-segmentation}: Ultrasound images for nerve structure segmentation, useful for ultrasound-guided regional anesthesia applications.
    
    \item \textbf{MuscleUS}~\cite{marzola2021deep}: Ultrasound images of muscle tissue with corresponding anatomical segmentations for musculoskeletal analysis.
    
    \item \textbf{MMOTU}~\cite{zhao2022mmotu}: Multi-Modality Ovarian Tumor Ultrasound dataset containing ultrasound images with ovarian tumor segmentations.
    
    \item \textbf{BraTS}~\cite{baid2021rsna}: Brain Tumor Segmentation Challenge dataset containing multi-parametric MRI scans (T1, T1ce, T2, FLAIR) with expert annotations of tumor subregions.
    
    \item \textbf{FLARE}~\cite{ma2022fast}: Fast and Low-resource semi-supervised Abdominal oRgan sEgmentation challenge dataset containing CT scans with multi-organ annotations.
    
    \item \textbf{ToothSeg}~\cite{silva2018automatic}: Dental CT and X-ray images with tooth and anatomical structure segmentations for dental analysis.
    
    \item \textbf{ISIC}~\cite{codella2018skin}: International Skin Imaging Collaboration dataset containing dermoscopic images of skin lesions with segmentation masks for melanoma detection.
    
    \item \textbf{PAXRay}~\cite{seibold2022detailed}: Chest X-ray dataset with detailed anatomical structure segmentations for thoracic analysis.
    
    \item \textbf{BBBC038}~\cite{caicedo2019nucleus}: Broad Bioimage Benchmark Collection dataset containing diverse microscopy images of cell nuclei across varied experimental conditions with instance segmentation masks.
\end{itemize}

\subsubsection{Unseen Datasets}

\begin{itemize}
    \item \textbf{CHAOS}~\cite{kavur2021chaos}: Combined Healthy Abdominal Organ Segmentation challenge dataset containing both CT and MRI scans with annotations for liver, kidneys, and spleen.
    
    \item \textbf{LungRads}~\cite{nam2024lung}: CT images with segmentation of lung cancer patients.
    
    \item \textbf{Kvasir-SEG}~\cite{jha2019kvasir}: Gastrointestinal polyp segmentation dataset containing colonoscopy images with pixel-wise polyp annotations, manually verified by experienced gastroenterologists.
    
    \item \textbf{CVC-ClinicDB}~\cite{bernal2015wm}: Colonoscopy image database for polyp detection and segmentation containing frames extracted from colonoscopy videos with ground truth polyp masks.
    
    \item \textbf{LGG}~\cite{buda2019association}: Low-Grade Glioma segmentation dataset containing brain MRI scans with FLAIR abnormality segmentation masks for tumor delineation.
    
\end{itemize}

\section{Qualitative Results}
\label{supp:sec:qualitative}

\begin{figure}[t]
\centering
\includegraphics[width=1.0\linewidth]{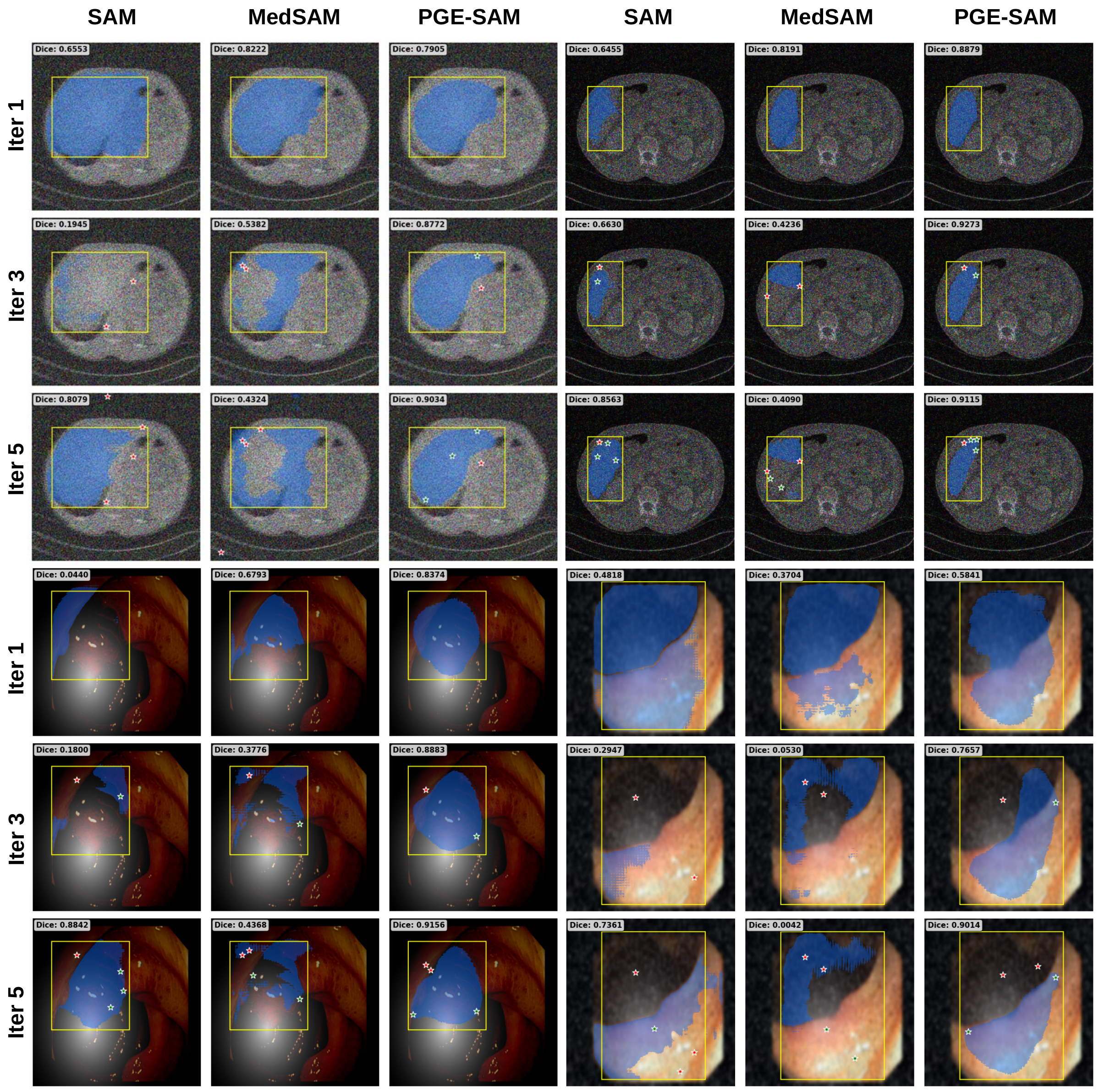}

\caption{\textbf{Qualitative results on \dataName{} (medical images).} Sample predictions under degraded conditions on \dataName{} dataset.}
\label{fig:chaos_quan}
\end{figure}

\begin{figure}[t]
\centering
\includegraphics[width=1.0\linewidth]{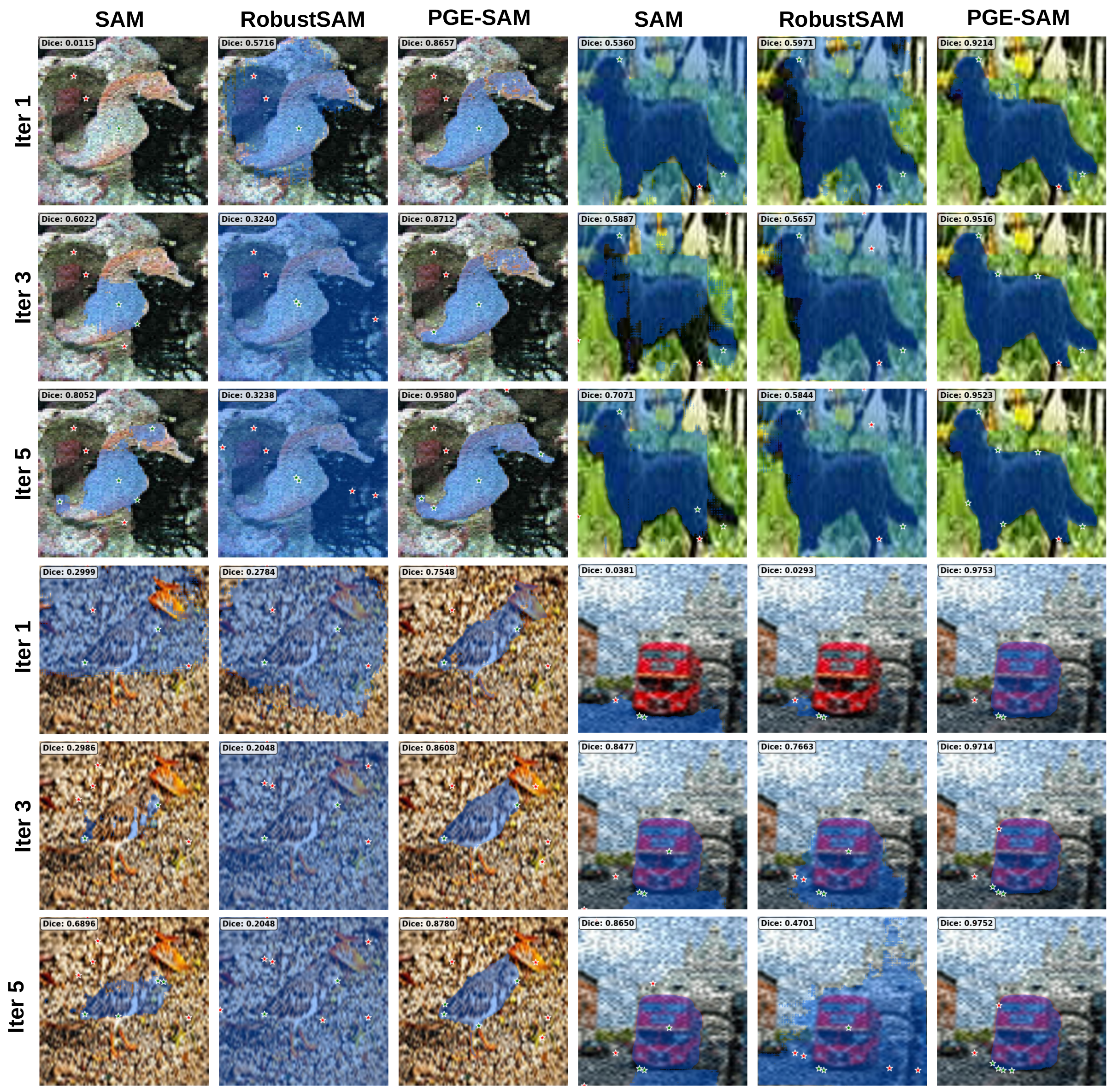}

\caption{\textbf{Qualitative results on LQ-Seg (natural images).} Sample predictions on the LQ-Seg dataset under synthetic degradation.}
\label{fig:lqseg_viz}
\end{figure}

\begin{figure}[t]
\centering

\includegraphics[width=1.0\linewidth]{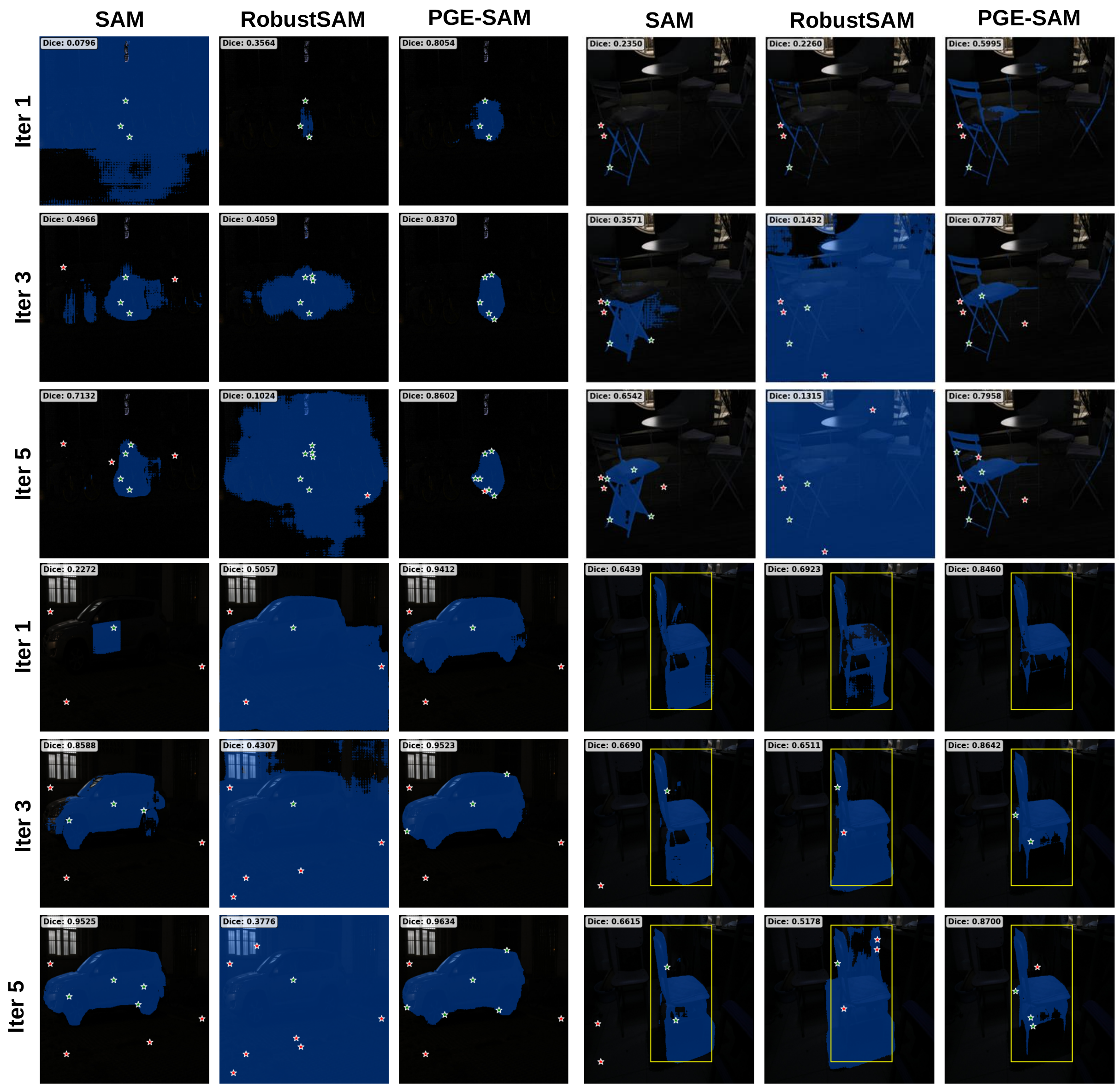}

\caption{\textbf{Qualitative results on real-world low-light images.} Sample predictions on the LIS dataset~\cite{lis}, which contains naturally occurring low-light conditions.}
\label{fig:lis_viz}
\end{figure}

\begin{figure}[t]
\centering

\includegraphics[width=1.0\linewidth]{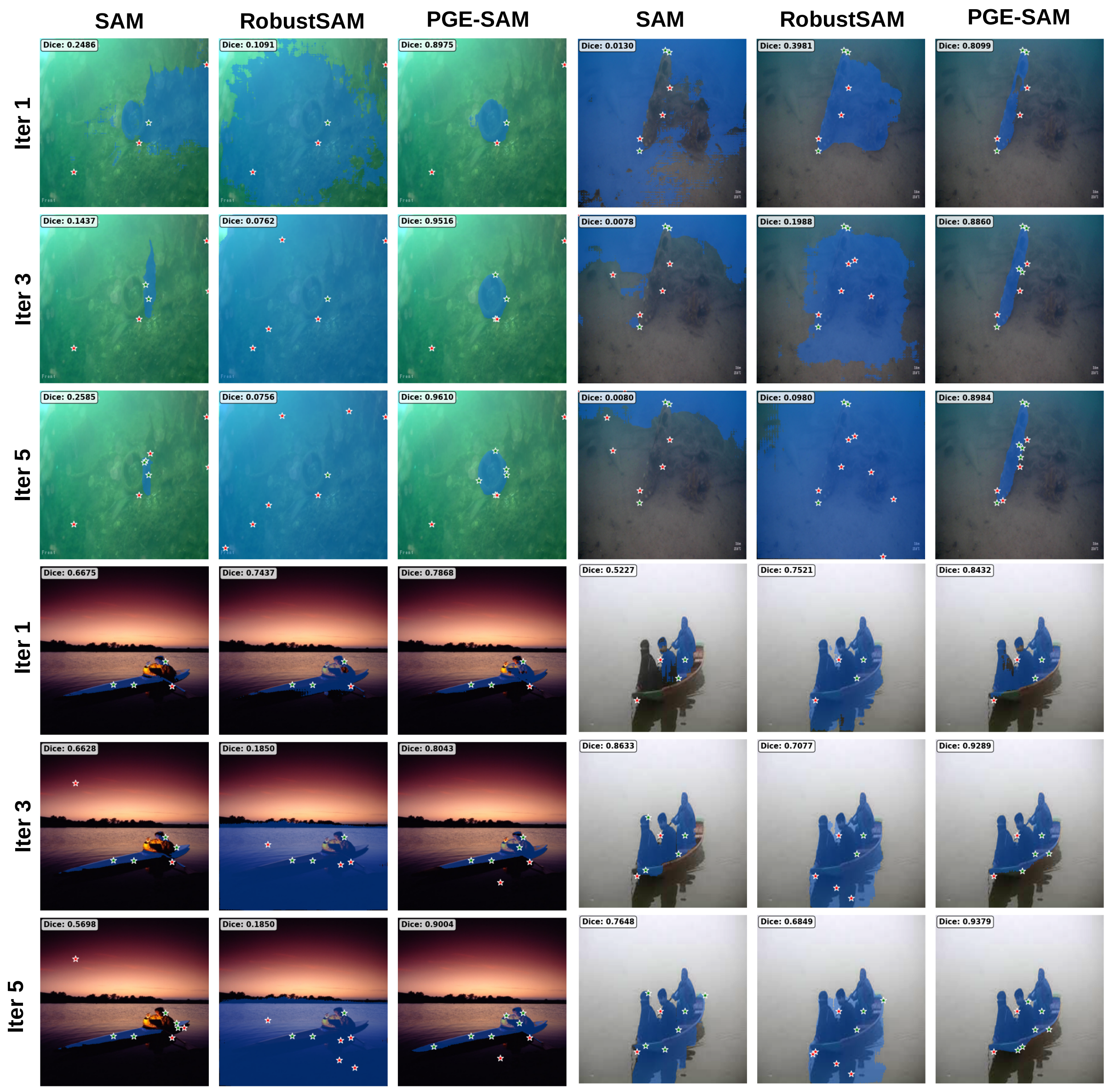}

\caption{\textbf{Qualitative results on real-world adverse conditions.} Sample predictions on the SMDD dataset~\cite{smdd} (underwater debris) and WXSDO dataset~\cite{wxsdo} (adverse weather).}
\label{fig:wxsdo_smdd_viz}
\end{figure}

We provide additional qualitative comparisons on medical images from \dataName{} (Fig.~\ref{fig:chaos_quan}), synthetic degradation on LQ-Seg (Fig.~\ref{fig:lqseg_viz}), and real-world degraded images (Figs.~\ref{fig:lis_viz} and~\ref{fig:wxsdo_smdd_viz}).

\section{Detailed Per-Step Iterative Results}
\label{supp:sec:detailed_iterative}

We present per-step IoU and Dice curves for each individual dataset across multiple evaluation settings. These visualizations complement the aggregated tables in the main paper by revealing the iterative dynamics of each method. Importantly, the figures demonstrate that \mName{} not only benefits from iterative refinement but also outperforms all baselines at the very first interaction step, confirming that its gains stem from the prompt-guided feature enhancement architecture rather than solely from training with iterative supervision.

\begin{figure}[tb]
\centering
\includegraphics[width=1.0\linewidth]{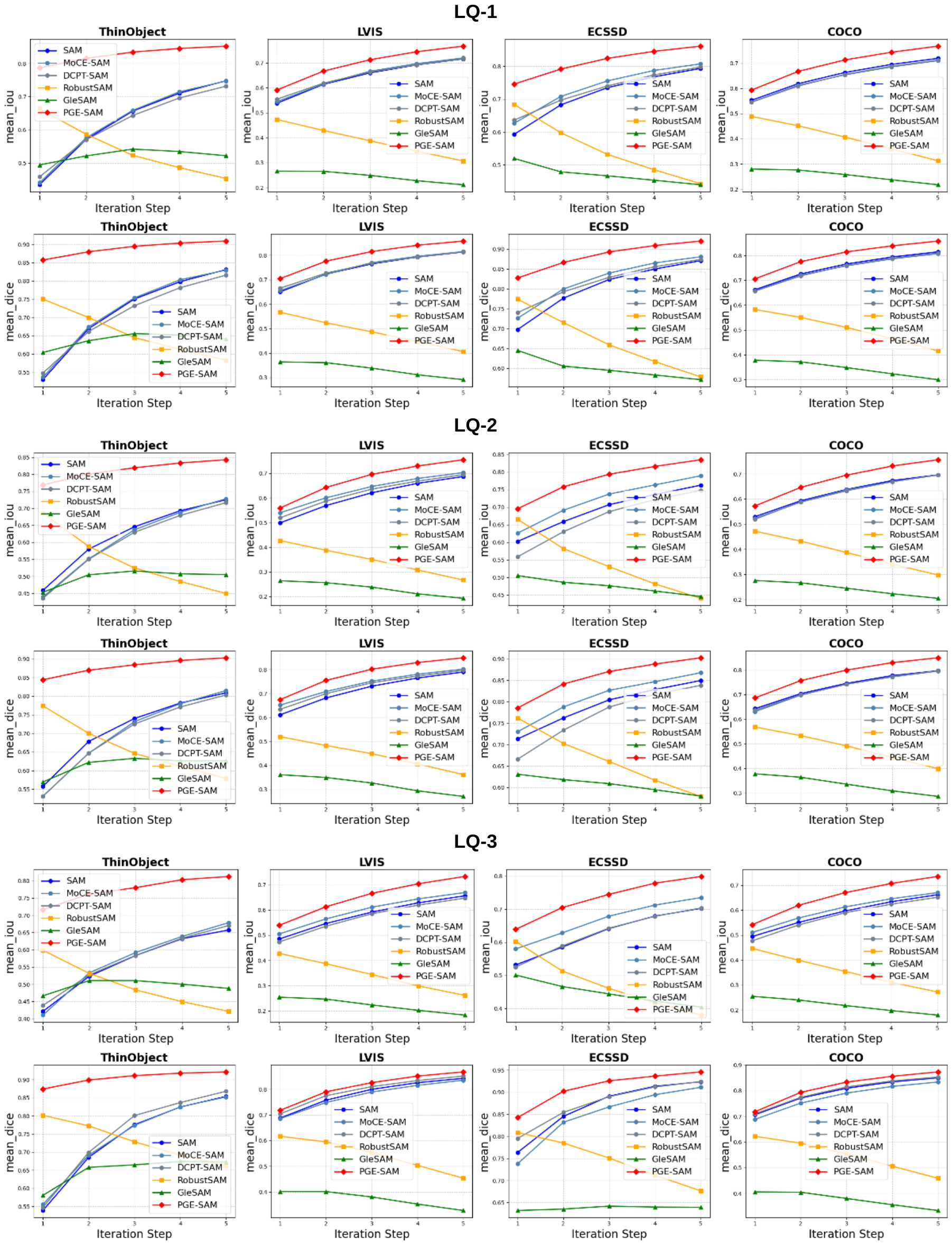}
   \caption{\textbf{Per-step IoU and Dice on LQ-Seg~\cite{glesam} across three degradation levels.} \mName{} outperforms all baselines from the first interaction step on both seen and unseen datasets.}
\label{fig:iter_lqseg}
\end{figure}

\paragraph{\textbf{LQ-Seg degradation.}} Fig.~\ref{fig:iter_lqseg} reports per-step performance across all three degradation levels (LQ-1, LQ-2, LQ-3) on both seen (ThinObject-5K, LVIS) and unseen (ECSSD, COCO) datasets. \mName{} achieves the highest IoU and Dice from the first step onward, and continues to improve steadily with each subsequent interaction. In contrast, RobustSAM and GleSAM either stagnate or degrade after early steps due to their lack of iterative awareness. Notably, even compared to SAM, which preserves the original iterative mechanism, \mName{} starts with a significantly higher initial prediction and maintains this advantage throughout, demonstrating the effectiveness of the Feature Enhancer in producing better initial representations.

\begin{figure}[tb]
\centering
\includegraphics[width=1.0\linewidth]{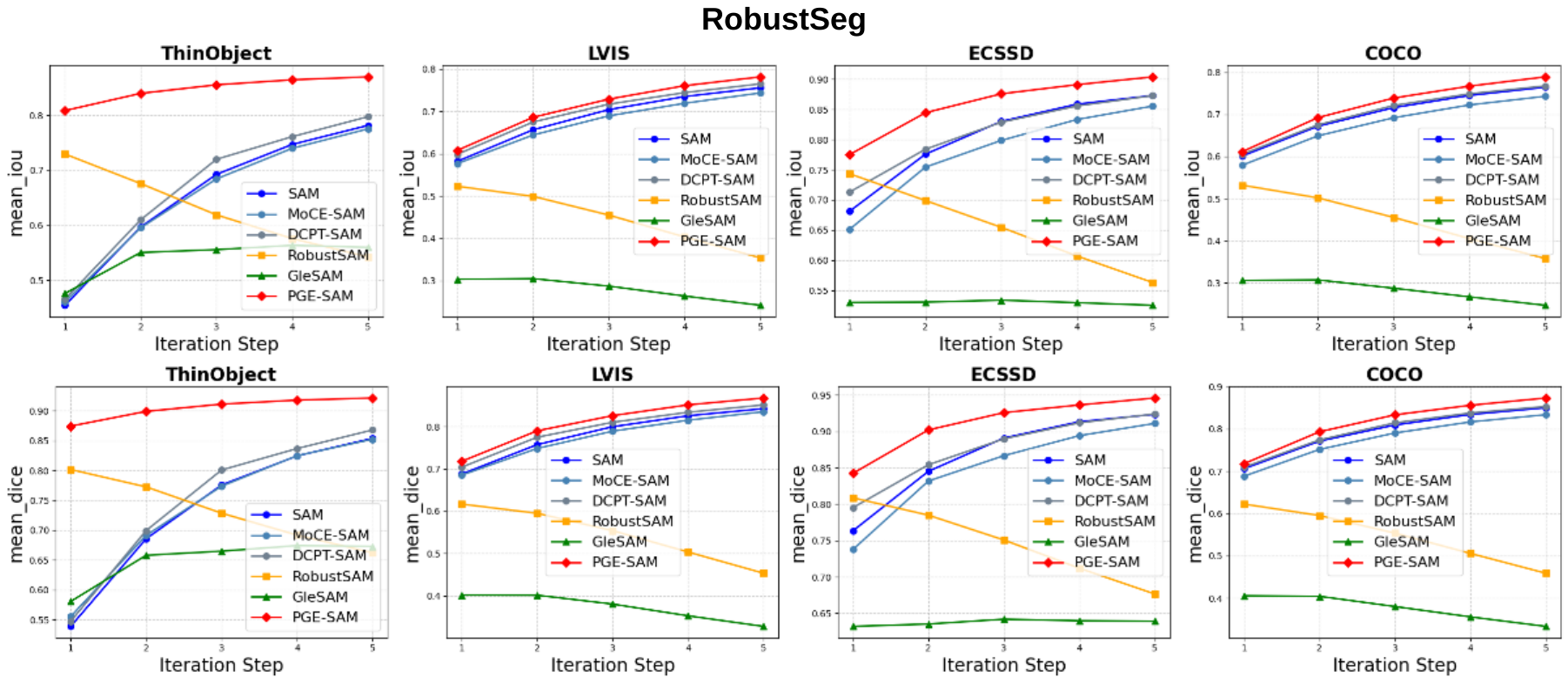}
   \caption{\textbf{Per-step IoU and Dice on RobustSeg-style degradation~\cite{robustsam}.} This degradation type is unseen during training. \mName{} generalizes effectively and leads from the first step.}
\label{fig:iter_robustseg}
\end{figure}

\paragraph{\textbf{RobustSeg-style degradation.}} Fig.~\ref{fig:iter_robustseg} shows iterative performance under the RobustSeg degradation protocol~\cite{robustsam}, a degradation type not seen during \mName{}'s training. Despite the zero-shot nature of this setting, the same pattern holds: \mName{} leads from the first step and iteratively improves, further validating the generalization of our approach to unseen degradation types.

\begin{figure}[tb]
\centering
\includegraphics[width=1.0\linewidth]{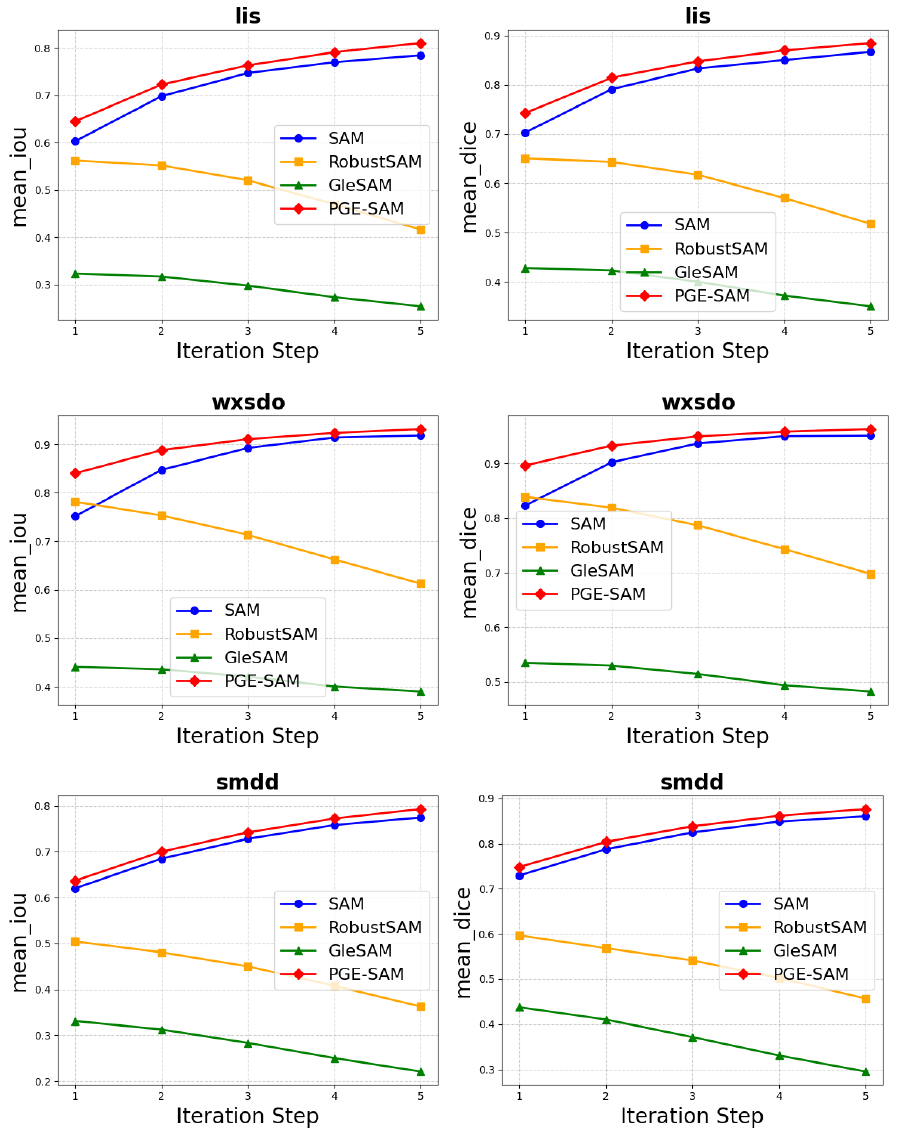}
   \caption{\textbf{Per-step IoU and Dice on real-world degradation datasets:} LIS~\cite{lis} (low-light), WXSDO~\cite{wxsdo} (adverse weather), and SMDD~\cite{smdd} (marine debris). All datasets are zero-shot with respect to \mName{}'s training.}
\label{fig:iter_real_world}
\end{figure}

\paragraph{\textbf{Real-world degradation.}} Fig.~\ref{fig:iter_real_world} presents results on three real-world adverse condition datasets: LIS~\cite{lis} (low-light), WXSDO~\cite{wxsdo} (adverse weather), and SMDD~\cite{smdd} (marine debris under murky water). These datasets feature authentic, non-synthetic degradation that our model has never encountered during training. \mName{} consistently outperforms SAM and other baselines across all three datasets from the first interaction and continues to outperform all baselines through the final iteration.

\begin{figure}[tb]
\centering
\includegraphics[width=1.0\linewidth]{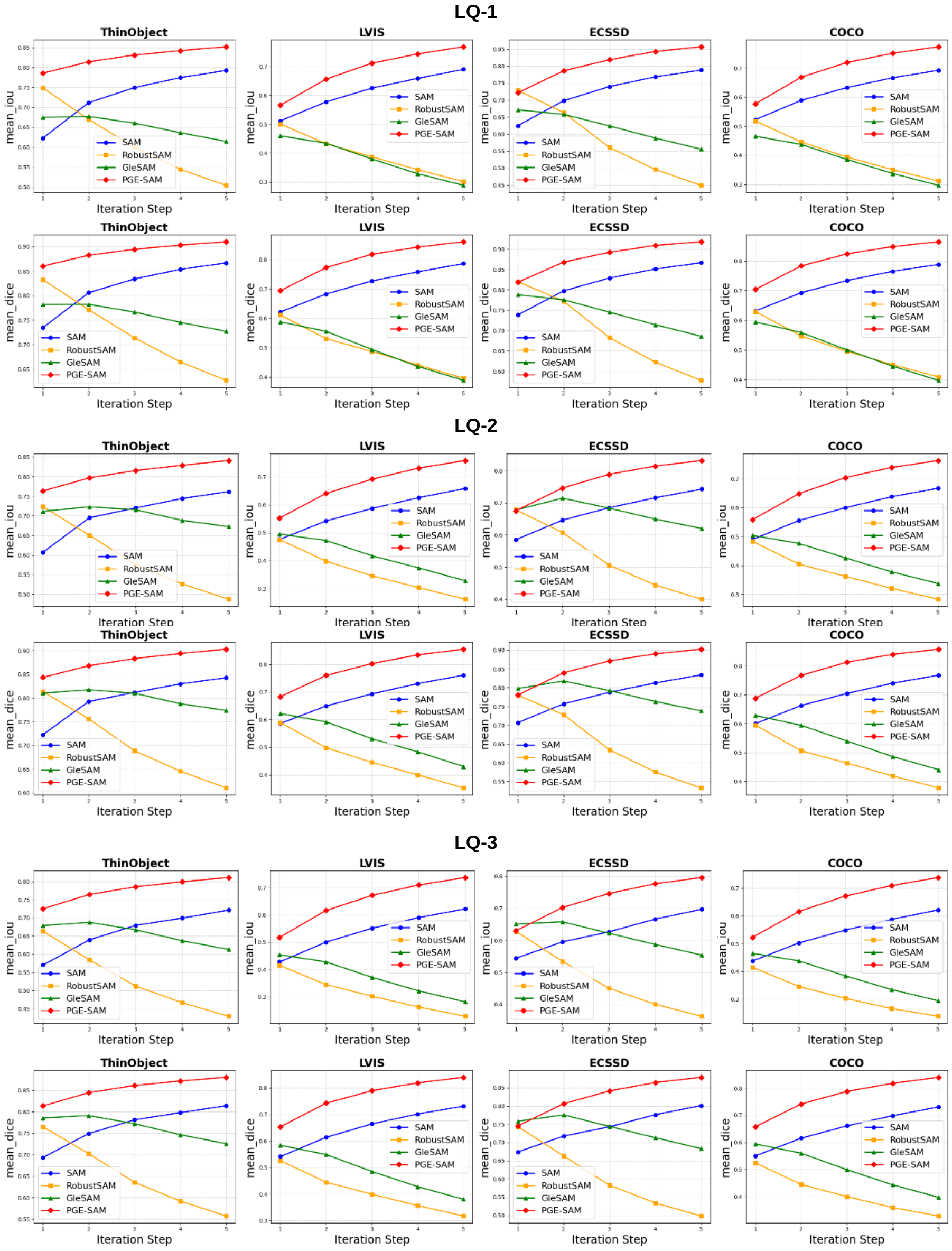}
   \caption{\textbf{Per-step IoU on LQ-Seg with 3 random positive point initialization.} \mName{} achieves the strongest overall iterative performance across all four datasets and degradation levels.}
\label{fig:3_point_lqseg}
\end{figure}

\begin{figure}[tb]
\centering
\includegraphics[width=1.0\linewidth]{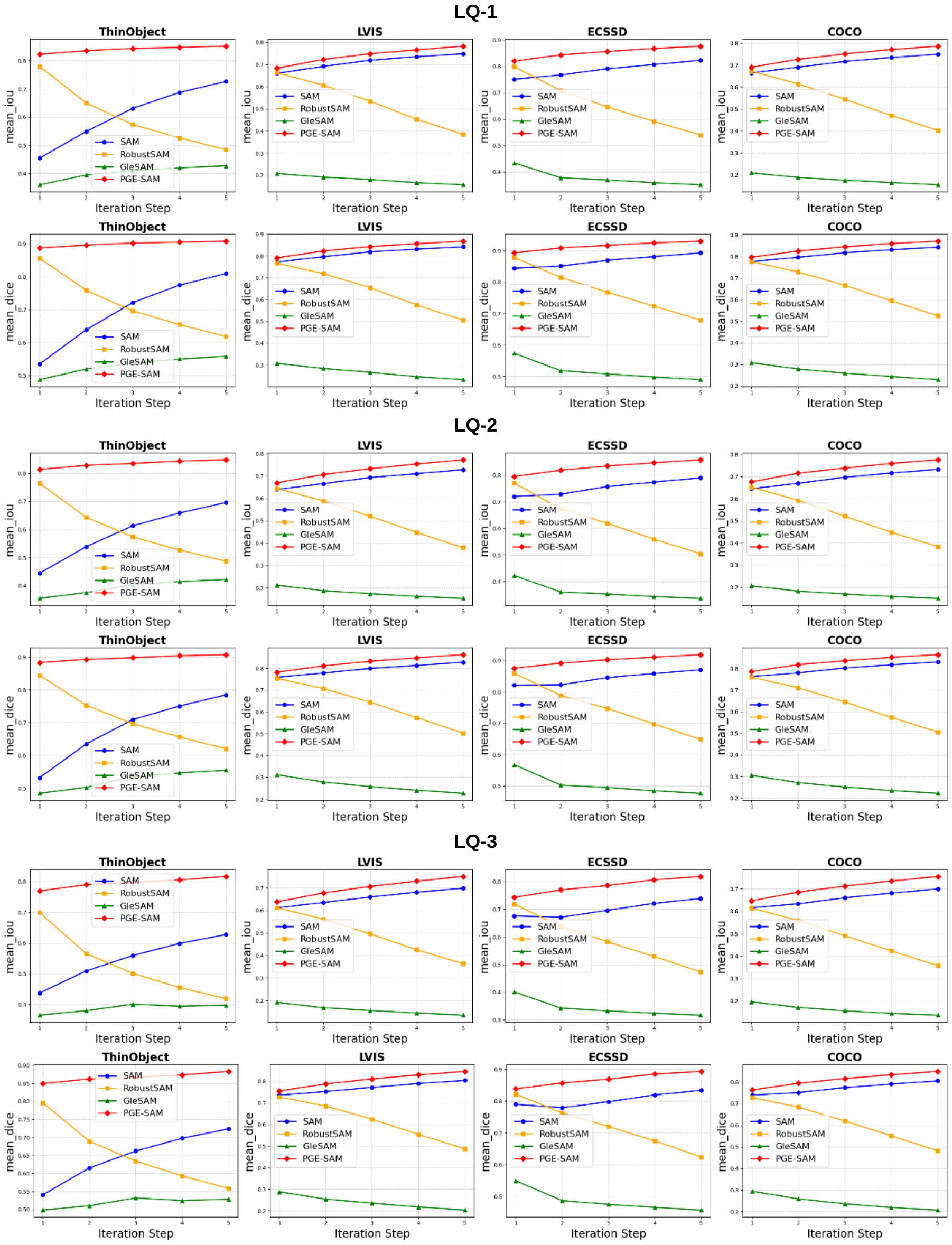}
   \caption{\textbf{Per-step IoU on LQ-Seg with box prompt initialization.} \mName{} maintains consistent performance, while GleSAM's performance collapses under this prompt type.}
\label{fig:box_lqseg}
\end{figure}

\paragraph{\textbf{Different prompt types.}} Figs.~\ref{fig:3_point_lqseg} and~\ref{fig:box_lqseg} show per-step performance when the initial prompt is fixed to either 3 random positive points or a bounding box, respectively. Under both prompt types, \mName{} consistently outperforms all baselines across all four LQ-Seg datasets when considering the full iterative trajectory. The iterative curves further corroborate the prompt sensitivity of GleSAM discussed in Sec.~\ref{supp:sec:expe}: while GleSAM performs comparably to SAM under 3-point initialization, its performance collapses under box prompts, whereas \mName{} remains robust regardless of prompt type.

\begin{figure}[tb]
\centering
\includegraphics[width=1.0\linewidth]{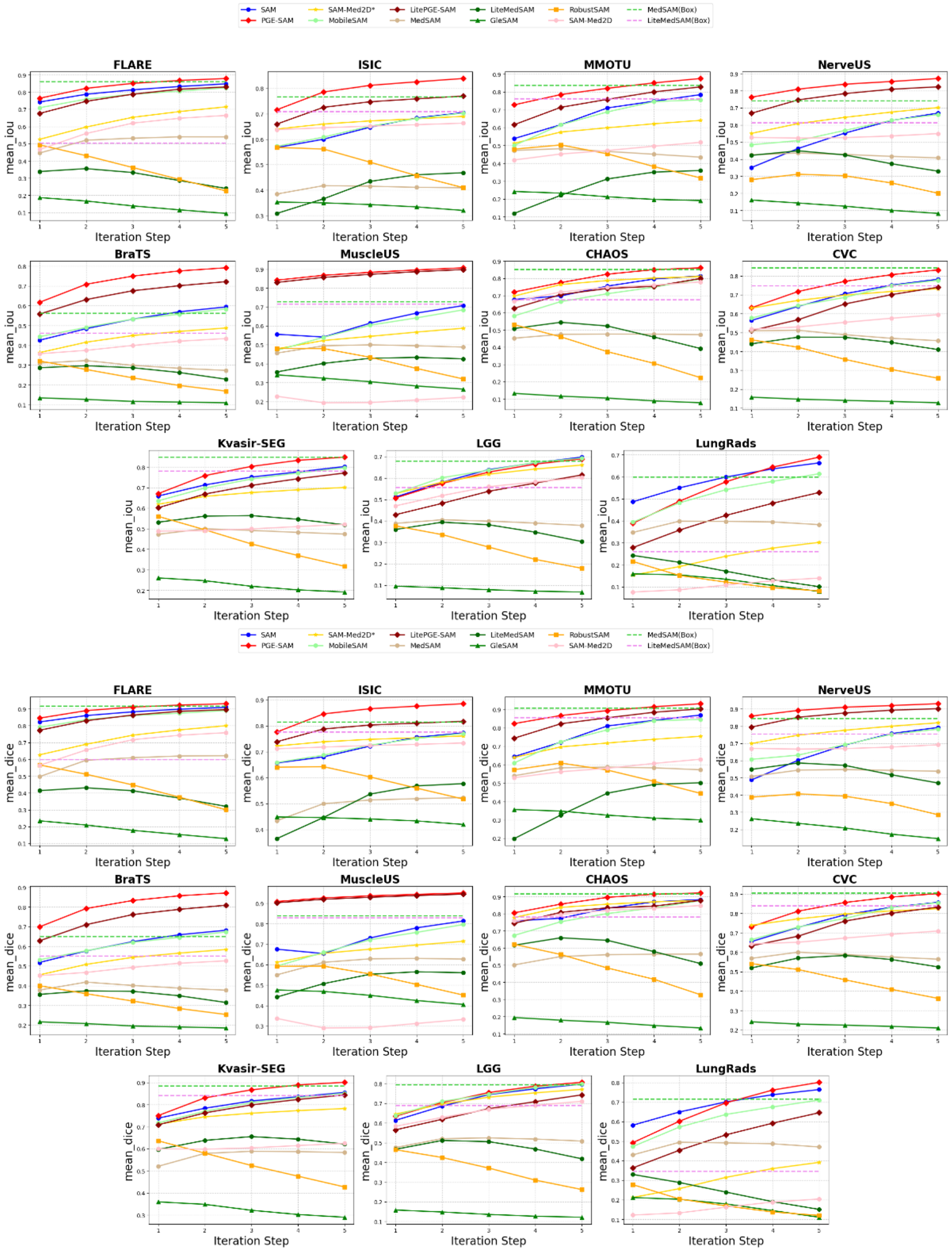}
   \caption{\textbf{Per-step IoU and Dice under LQ-1 degradation on \dataName{}.} Results shown for each dataset in \dataName{} validation set.}
\label{fig:med_lq1}
\end{figure}

\begin{figure}[tb]
\centering
\includegraphics[width=1.0\linewidth]{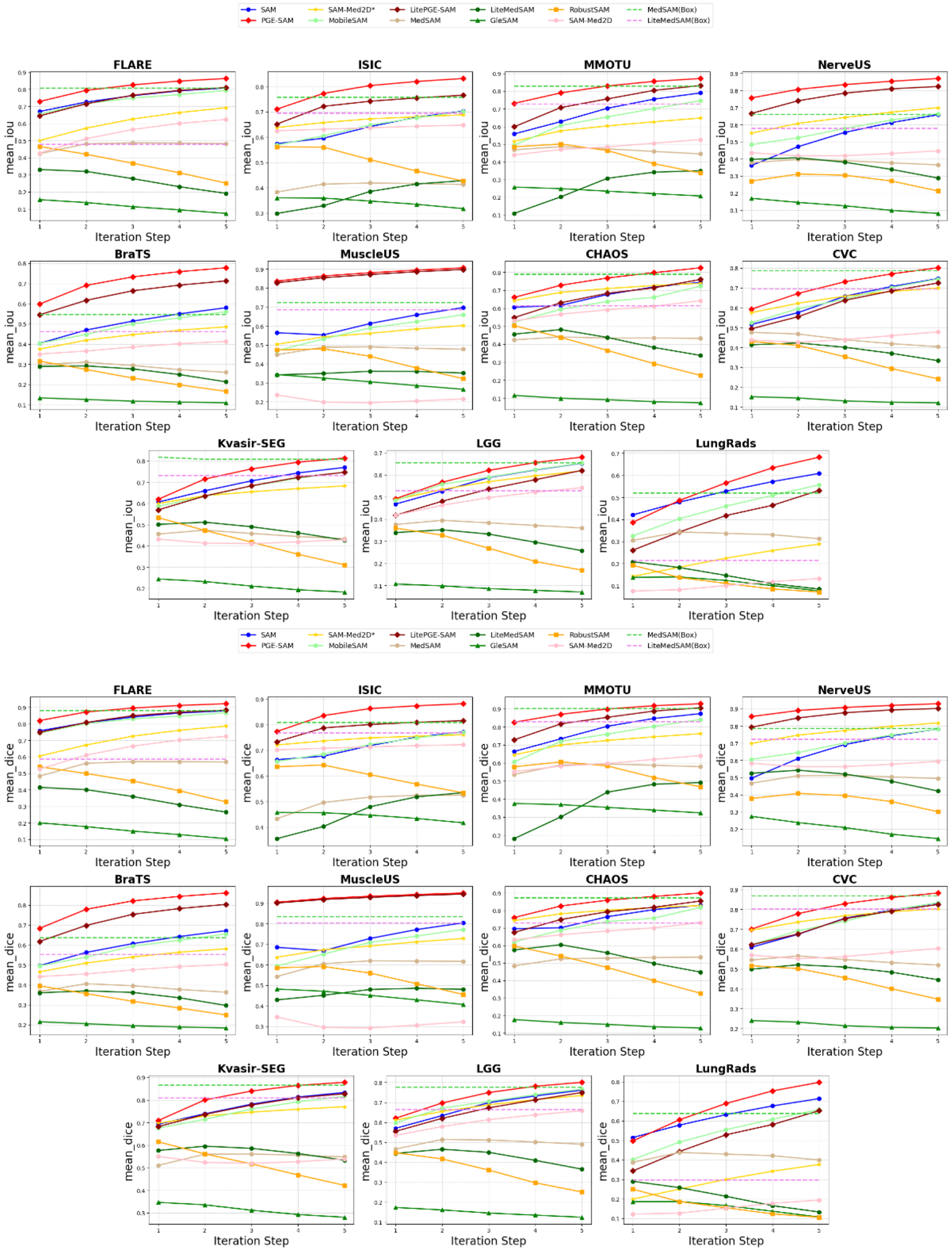}
   \caption{\textbf{Per-step IoU and Dice under LQ-2 degradation on \dataName{}.} Results shown for each dataset in \dataName{} validation set.}
\label{fig:med_lq2}
\end{figure}

\begin{figure}[tb]
\centering
\includegraphics[width=1.0\linewidth]{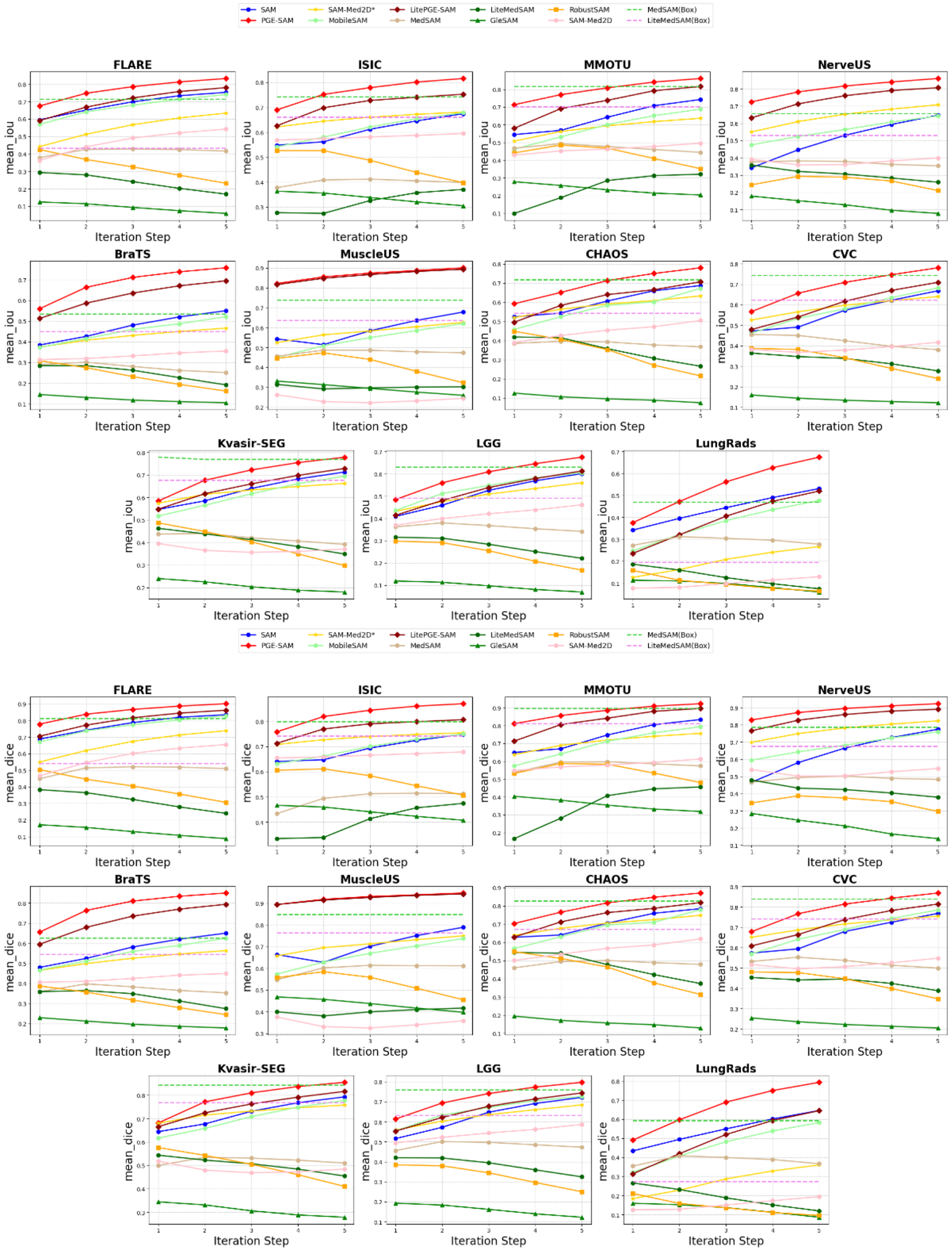}
   \caption{\textbf{Per-step IoU and Dice under LQ-3 degradation on \dataName{}.} Results shown for each dataset in \dataName{} validation set.}
\label{fig:med_lq3}
\end{figure}

\begin{figure}[tb]
\centering
\includegraphics[width=1.0\linewidth]{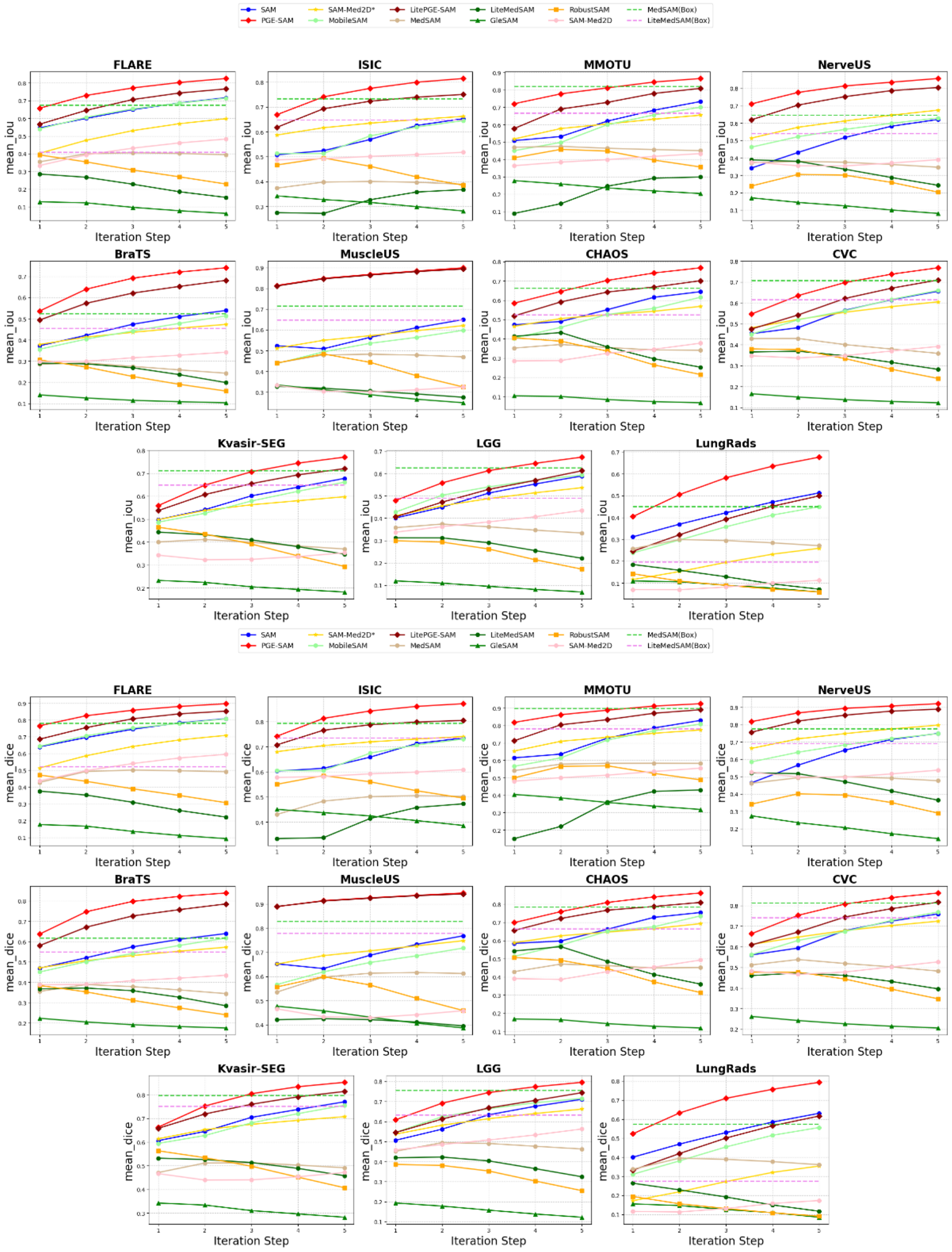}
   \caption{\textbf{Per-step IoU and Dice under LQ-3+ degradation on \dataName{}.} Results shown for each dataset in \dataName{} validation set.}
\label{fig:med_lq3+}
\end{figure}

\paragraph{\textbf{Medical image domain.}} Figs.~\ref{fig:med_lq1}--\ref{fig:med_lq3+} present per-step results on each dataset within the \dataName{} validation set for all degradation levels. Overall, \mName{} consistently achieves the best performance across both seen and unseen splits.

\clearpage
\bibliographystyle{splncs04}
\bibliography{main}
\end{document}